\documentclass[5p,times,twocolumn]{elsarticle}

\usepackage{amsmath,amssymb,amsfonts,color}
\usepackage{graphicx}
\usepackage{textcomp}
\usepackage[dvipsnames]{xcolor}
\usepackage{enumerate}
\usepackage{setspace}
\usepackage{bm}
\usepackage{comment}
\usepackage{caption}
\usepackage{tipa}
\usepackage{footnote}
\usepackage{subfigure}
\usepackage{algorithm}
\usepackage{algpseudocode}

\usepackage{multirow}
\usepackage{multicol}
\usepackage{subfigure}
\usepackage{multirow}
\usepackage{url}
\usepackage{booktabs}
\usepackage{tabularx}
\usepackage{adjustbox}
\usepackage{arydshln}

\usepackage{hyperref}
\begin{document}

\begin{frontmatter}

\title{Cross-domain Multi-step Thinking: Zero-shot Fine-grained \\Traffic Sign Recognition in the Wild}
\author{Yaozong Gan${}^\text{a}$}
\ead{gan@lmd.ist.hokudai.ac.jp}
\author{Guang Li${}^\text{b}$}
\ead{guang@lmd.ist.hokudai.ac.jp}
\author{Ren Togo${}^\text{c}$}
\ead{togo@lmd.ist.hokudai.ac.jp}
\author{Keisuke Maeda${}^\text{c}$}
\ead{maeda@lmd.ist.hokudai.ac.jp}
\author{Takahiro Ogawa${}^\text{c}$}
\ead{ogawa@lmd.ist.hokudai.ac.jp}
\author{Miki Haseyama${}^\text{c}$}
\ead{mhaseyama@lmd.ist.hokudai.ac.jp}
\address{${}^\text{a}$Graduate School of Information Science and Technology, Hokkaido University, \\
           N-14, W-9, Kita-Ku, Sapporo, 060-0814, Japan}
\address{${}^\text{b}$Education and Research Center for Mathematical and Data Science, Hokkaido University, \\
           N-12, W-7, Kita-Ku, Sapporo, 060-0812, Japan}
\address{${}^\text{c}$Faculty of Information Science and Technology, Hokkaido University, Hokkaido University, \\
           N-14, W-9, Kita-Ku, Sapporo, 060-0814, Japan}

\begin{abstract}
In this study, we propose \textbf{C}ross-\textbf{d}omain \textbf{M}ulti-step \textbf{T}hinking (\textbf{CdMT}) to improve zero-shot fine-grained traffic sign recognition (TSR) performance in the wild. Zero-shot fine-grained TSR in the wild is challenging due to the cross-domain problem between clean template traffic signs and real-world counterparts, and existing approaches particularly struggle with cross-country TSR scenarios, where traffic signs typically differ between countries. The proposed CdMT framework tackles these challenges by leveraging the multi-step reasoning capabilities of large multimodal models (LMMs). We introduce context, characteristic, and differential descriptions to design multiple thinking processes for  LMMs. Context descriptions, which are enhanced by center coordinate prompt optimization, enable the precise localization of target traffic signs in complex road images and filter irrelevant responses via novel prior traffic sign hypotheses. Characteristic descriptions, which are derived from in-context learning with template traffic signs, bridge cross-domain gaps and enhance fine-grained TSR. Differential descriptions refine the multimodal reasoning ability of LMMs by distinguishing subtle differences among similar signs. CdMT is independent of training data and requires only simple and uniform instructions, enabling it to achieve cross-country TSR.~\textcolor{black}{We conducted extensive experiments on three benchmark datasets and two real-world datasets from different countries. The proposed CdMT framework achieved superior performance compared with other state-of-the-art methods on all five datasets, with recognition accuracies of 0.93, 0.89, 0.97, 0.89, and 0.85 on the GTSRB,  BTSD, TT-100K, Sapporo, and Yokohama datasets, respectively.}
\end{abstract}


\end{frontmatter}

\section{Introduction}
\par Ensuring traffic safety remains a critical issue in the real world~\cite{liu2017provid}. The latest statistics from the World Health Organization show that road traffic injuries are the leading cause of death among children and adolescents aged 5--29 years and that approximately 1.19 million people die each year due to road traffic accidents~\footnote{https://www.who.int/news-room/fact-sheets/detail/road-traffic-injuries}.
Furthermore, road traffic accidents lead to substantial economic losses and impose a significant burden on society~\cite{tan2020cost}. Consequently, there is an urgent need to reduce the number of road traffic accidents.
\par Traffic sign recognition (TSR) enables vehicles to identify traffic signs on dynamic road scenes. As an important part of the road, it is crucial to effectively recognize traffic signs for traffic safety. Advanced driver assistance systems help drivers make safer decisions by evaluating driving conditions based on traffic sign data and alerting drivers to road inconsistencies~\cite {romdhane2016improved}. In addition, TSR helps global positioning systems and map service providers update their databases. Consequently, TSR technology has attracted widespread attention.

\begin{figure}[t]
        \centering
        \includegraphics[width=8.5cm]{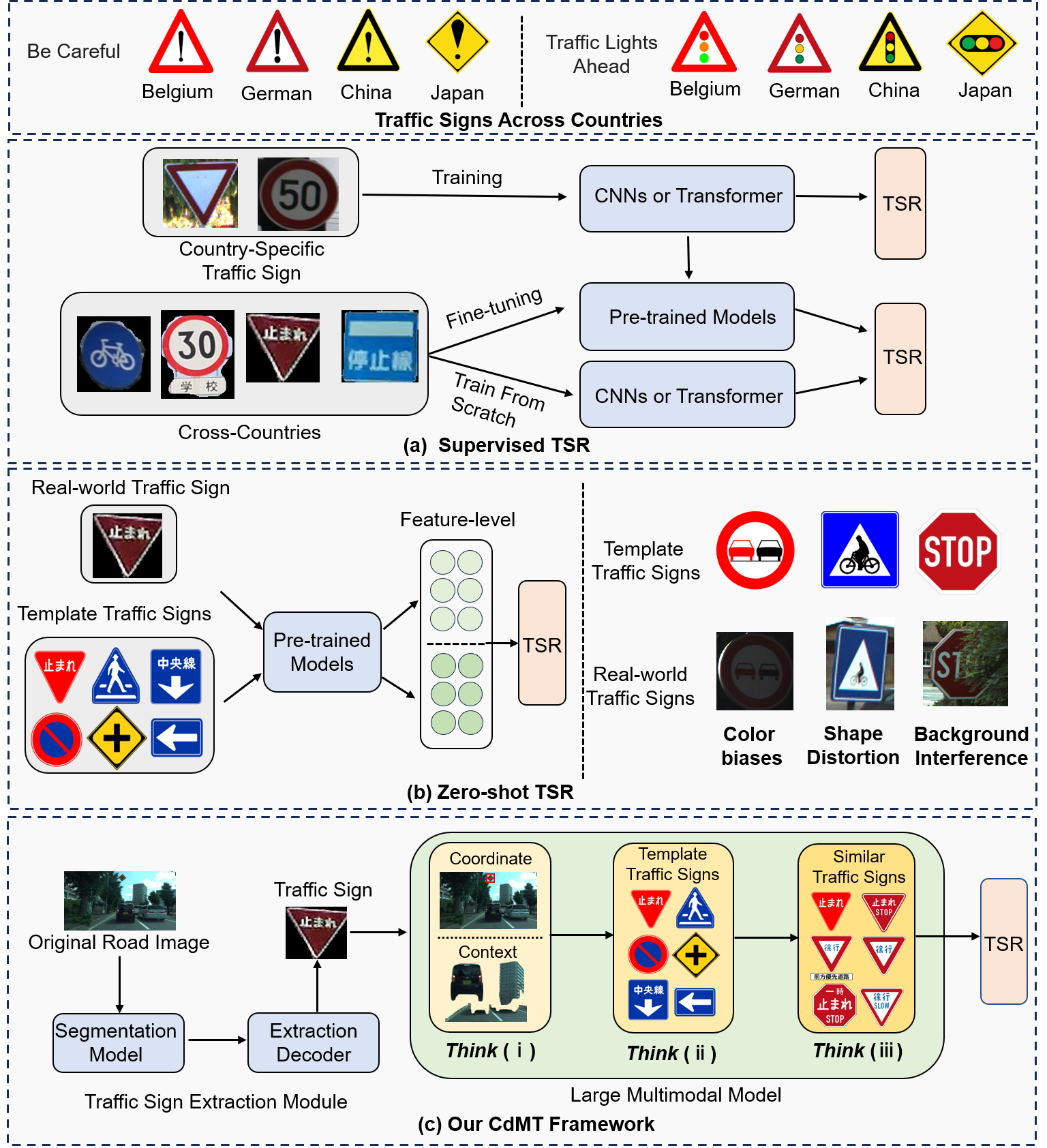}
        \caption{Comparison of different TSR methods. (a) Supervised TSR, which requires a large amount of training data and fine-tuning. (b) Feature-level TSR, which requires no training data. Cross-domain differences exist between target and template traffic signs. (c) Our CdMT framework, which stimulates the multi-thinking capabilities of large multimodal models (LMMs) without requiring training data.}
        \label{fig1}
\end{figure}
\par Early TSR studies focus on using hand-crafted features such as the histogram of oriented gradients (HOG)~\cite{yucong2021traffic,zaklouta2011segmentation,zaklouta2011real,huang2016efficient} and scale-invariant feature transform (SIFT)~\cite{lowe2004distinctive,yin2015fast,malik2014detection}. Newer methods are based on convolutional neural networks (CNNs) ~\cite{guo2023grtr, bi2021improved, baruah2023traffic, luo2017traffic} and vision transformers~\cite{manzari2022pyramid,luo2023pre,guo2024signparser,zheng2022evaluation}, which use the feature representation capabilities of convolutional layers or attention mechanisms to perform supervised recognition on country-specific traffic sign images as shown in Fig.~\ref{fig1}-(a). These methods have two major limitations: first, the supervised feature learning process requires a large amount of carefully crafted training data for traffic signs, which is usually clearly visible. In contrast, traffic signs on real-world roads can be blurred or broken due to the influence of dynamic road scenes, weather, and other factors.
Second, to unify the traffic signs across countries, the Vienna Convention on Road Traffic~\cite{economic1968convention} stipulates more than 300 different traffic sign categories; however, only 83 countries have signed the treaty. Thus, traffic signs vary significantly across most countries. In addition, some visual differences still exist between traffic signs in countries that have signed the treaty. As shown at the top of Fig.~\ref{fig1}, even for the same type of traffic signs ``Be Careful'' and ``Traffic Lights Ahead,'' differences exist between countries. Because they are trained on country-specific datasets, these methods require fine-tuning or training from scratch when recognizing traffic signs in other countries. These are costly due to data policy restrictions in various countries and the difficulty in obtaining data in underdeveloped regions. Some methods based on unsupervised learning or feature matching have been proposed to solve the cross-country TSR problem~\cite{ren2009general, supriyanto2016unsupervised, zhou2020few, gan2023zero, 1520577215790999680}. These approaches typically employ zero-shot learning strategies, reducing reliance on extensive training data and addressing the applicability challenges of cross-country TSR. However, as shown in Fig.~\ref{fig1}-(b), significant cross-domain discrepancies exist between target and template traffic signs. In real-world scenarios, traffic signs may exhibit color biases or shape distortions and are typically embedded in complex environments such as roads or streets, which are typically partially occluded by objects such as trees, billboards, pedestrians, or vehicles. In contrast, template traffic sign images are standardized in color and appearance and are free from background interference. Thus, performing pairwise matching at the feature level tends to amplify these discrepancies, thereby limiting the recognition accuracy of existing methods.

\par Recent breakthroughs in large language models (LLMs)~\cite{brown2020language,chowdhery2023palm,openai2024gpt4,touvron2023llama} have introduced general artificial intelligence models that can solve various complex natural language tasks, many of which are approaching the performance level of human experts~\cite{openai2024gpt4, bubeck2023sparks}. In addition to text, other modalities, including images, are used in the real world. Many studies have proposed several visual-text LMMs~\cite{tsimpoukelli2021multimodal,touvron2023llama,openaisystem1,openaisystem,openaitech} to solve various visual problems existing in the real world~\cite{yunxinli,ghali2025enhancing,zeng2025kosel,dai2025large,gan2024cross,cook2024llm}. In traffic safety, LMMs exhibit significant application value in constructing future intelligent transportation systems~\cite {zheng2023chat}. Furthermore, LMMs have significant potential in autonomous driving and can revolutionize the conventional human-vehicle interaction model~\cite{cui2024survey}. Users can communicate requests through languages, gestures, and even eyes, and LMMs provide real-time in-vehicle feedback through integrated visual displays. However, despite the unprecedented recognition capabilities of LMMs, their research in TSR is limited. In general tasks, raw images are typically directly input into the LMM for recognition. On the one hand, it is difficult to recognize traffic signs directly as they are too small, e.g., in a road image with 1,280~$\times$~960 pixels, the traffic sign may be only 30~$\times$~30 pixels. On the other hand, unlike recognizing objects such as ``cats'' and ``dogs,'' different types of traffic signs are highly similar and TSR requires accurate recognition at a fine-grained level. Therefore, detailed studies are required to stimulate the potential of LMMs to realize fine-grained TSR.
 
\par In this study, we propose \textbf{C}ross-\textbf{d}omain \textbf{M}ulti-step \textbf{T}hinking (\textbf{CdMT}), a novel framework to tackle the current challenges of zero-shot fine-grained TSR in real-world settings. Unlike conventional methods that struggle with cross-domain disparities between standardized template traffic signs and their real-world counterparts, CdMT uses LMMs to achieve robust TSR without dependence on specific training data. As shown in Fig.~\ref{fig1}-(c), the proposed method begins with a traffic sign extraction module that locates and extracts traffic signs in the original road image while excluding potential background interference. To stimulate the recognition ability of the LMM, multiple thinking processes are designed to inspire the LMM to improve fine-grained TSR.
\newline\textit{\textbf{Think}} (\textbf{\romannumeral1}): As previously mentioned, recognizing traffic signs directly from original road images is inherently difficult due to their small size and contextual ambiguity. We propose context descriptions that contain important contextual information about traffic signs, such as crosswalks, vehicles, and pedestrians. Referencing the real-world question-answering and prompting process, we elaborate on a prompting strategy that allows the LMM to generate context descriptions while giving potential candidate answer options, named the prior traffic sign hypothesis. The prior traffic sign hypothesis helps filter irrelevant answers and reduce the difficulty of subsequent thinking. To handle images with multiple traffic signs, we introduce a center coordinate-based optimization, which enables the LMM to swiftly pinpoint the target sign and produce accurate descriptions, thereby overcoming the limitations of unfocused global analysis.
\newline\textit{\textbf{Think}} (\textbf{\romannumeral2}): Fine-grained TSR demands precise classification beyond coarse feature detection, a task in which LMMs typically struggle because of limited domain-specific knowledge. We address this problem by introducing in-context learning with template traffic signs. Specifically, considering the three important characteristics of traffic signs, namely, shape, color, and composition, we generate the characteristic description of each type of template traffic sign via in-context learning. The characteristic descriptions stimulate the fine-grained perceptual ability of the LMM. The template traffic signs can be easily obtained from the traffic sign databases, ensuring practicality and scalability across regions.
\newline\textit{\textbf{Think}} (\textbf{\romannumeral3}): The characteristics of certain types of traffic signs are highly similar, and our preliminary experiments demonstrate the limited ability of LMMs to recognize similar traffic signs. Therefore, we propose differential descriptions to emphasize the subtle dissimilarity between these traffic signs. Differential descriptions can further optimize the proposed strategy and improve the fine-grained recognition performance of the LMM.
\par During recognition, the LMM performs multiple thinking based on the generated descriptions. Our thinking strategy can largely motivate the LMM for fine-grained TSR. The proposed method is independent of training data and is applicable to cross-country TSR. In addition, the generation of each description is performed only once and requires only simple and uniform instructions. Our key contributions can be summarized as follows.
\begin{itemize}
    \item We propose the {\textbf{CdMT}} framework to stimulate the perceptual potential of fine-grained TSR by enhancing the multi-thinking ability of LMMs.
    \item We introduce the context descriptions of the original road images and propose the prior traffic sign hypothesis and center coordinate prompt optimization for localizing the target traffic sign in original road images containing multiple traffic signs and filtering irrelevant answers.
    \item We introduce in-context learning with template traffic signs, which enhances the fine-grained perceptual ability of LMMs. The characteristic descriptions reduce the cross-domain differences between the template and target traffic signs. We also generate differential description texts between similar traffic signs to optimize the multimodal thinking capability of the LMM.
    \item We conduct extensive experiments on three benchmark datasets and two real-world datasets from different countries, and CdMT achieves promising TSR results across all datasets.
\end{itemize}

\textcolor{black}{The remainder of this paper is organized as follows. Section 2 reviews related work on TSR and LMMs. Section 3 introduces the proposed CdMT framework in detail. Section 4 describes the experimental settings and presents the experimental results. Section 5 analyzes the limitations and discusses potential future works. Finally, Section 6 concludes the study.}
\section{Related Work}
\subsection{Traffic Sign Recognition}
\par TSR has become an extensively researched field, and many TSR approaches have been proposed. TSR is generally divided into two key steps: traffic sign detection (TSD) and traffic sign classification (TSC). TSD involves the localization and detection of traffic signs in road images, whereas TSC consists of the classification of the detected traffic signs. Many studies have applied conventional and deep learning methods to TSR.
\subsubsection{\textbf{Conventional TSR methods}}
\par Early TSR studies focused on performing recognition based on hand-crafted features and machine learning algorithms. For example, hand-crafted features are used to extract features from traffic signs, and machine learning algorithms are used to recognize the extracted features. Zaklouta et al.~\cite{zaklouta2011segmentation} introduced a real-time system for detecting and classifying circular and triangular traffic signs. Kus et al.~\cite{kus2008traffic} introduced a method for detecting and recognizing traffic signs by improving the SIFT~\cite{lowe2004distinctive} algorithm. The researchers enhanced SIFT by integrating features associated with the color of local regions. Huang et al.~\cite{huang2016efficient} proposed a streamlined TSR method by using HOG features and a single classifier trained using the extreme learning machine algorithm. HOG features strike a balance between redundancy and local details, improving the representation of distinctive shapes. Therefore, conventional methods rely heavily on hand-crafted features, which are sensitive to variations in lighting, occlusion, and complex backgrounds~\cite{kerim2021recognition}.
\begin{figure*}[htbp]{
    \centering
    \includegraphics[width=1\textwidth]{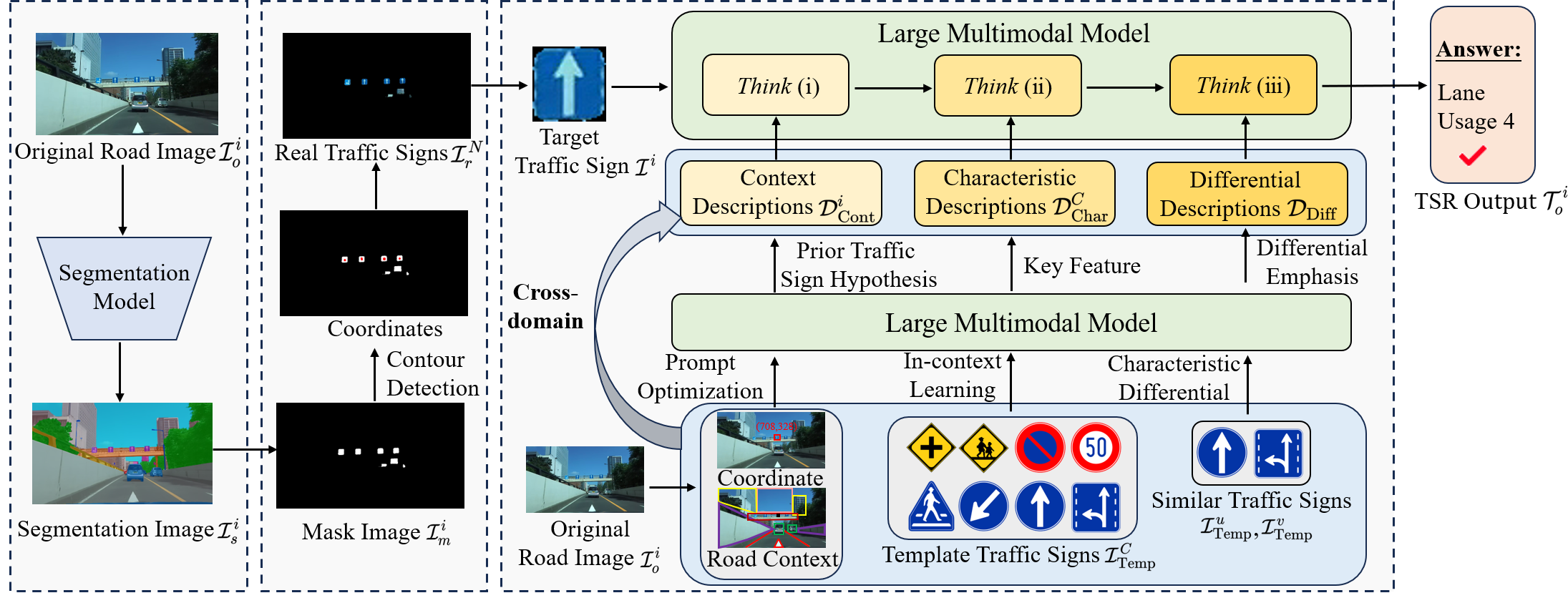}
    \caption{Overview of proposed method. The designed general network extracts traffic signs and performs multiple thinking processes for fine-grained TSR. \label{fig2}}}
\end{figure*}
\subsubsection{\textbf{Deep learning-based TSR methods}}
The emergence of deep learning has inspired TSR research. Compared with conventional hand-crafted feature-based methods, deep learning-based methods can better learn traffic sign image features.  Zhang et al.~\cite{zhang2020lightweight} introduced two lightweight networks for improving recognition accuracy with fewer parameters.  
Abudhagir et al.~\cite{abudhagir2022highly} used the LeNet model for TRS. Their CNN architecture comprised the first two layers adapted from LeNet, followed by two additional convolutional layers, a dropout layer, and a flattened layer. Zhu et al.~\cite{zhu2022traffic} proposed a TSR method based on YOLOv5. In addition, transformer-based TSR methods have been proposed. Zheng et al.~\cite{zheng2022evaluation} used a vision transformer (ViT)~\cite{dosovitskiy2020image} to perform a detailed TSR evaluation. 
Luo et al.~\cite{luo2023pre} proposed a TSR approach comprising a lightweight pre-locator network and a refined classification network based on Swin-Transformer~\cite{liu2021swin}. The pre-locator network identifies traffic sign sub-regions, and the refined classification network handles recognition within these regions.
Guo et al.~\cite{guo2024signparser} proposed an end-to-end framework that integrates component detection, content reasoning, and semantic description generation for understanding traffic signs. However, these supervised methods require fine-tuning or training from scratch when recognizing traffic signs in other
countries because they are trained on country-specific datasets. Nevertheless, TSR approaches have been introduced to solve this problem. For example, Cao et al.~\cite{cao2022sustainable} proposed a zero-shot method that synthesizes a hybrid feature representation by extracting both general and principal visual features from traffic sign images. Gan et al.~\cite{gan2023zero} introduced a zero-shot approach that uses midlevel features extracted from CNNs. However, because of the existence of cross-domain biases and the need for improving accuracy, more effective methods are expected to be explored.
\subsection{ Large Multimodal Models}
LLMs have received significant attention recently ~\cite{chang2023survey}. As demonstrated by existing work ~\cite{bubeck2023sparks}, LLMs can handle various tasks in contrast to previous models that are restricted to solving specific tasks. In addition, LMMs have been proposed~\cite{touvron2023llama,openaisystem1, openaisystem,team2023gemini,you2023ferret,rasheed2023glamm} to solve various visual problems in the real world. LMMs extend the capabilities of language models by integrating visual information as part of the input. This integration of visual data enables LMMs to efficiently understand and generate responses that contain both textual and visual prompts, thereby enabling richer context conversations in multimodal environments. In recent months, LMMs have also drawn attention in intelligent transportation applications, such as autonomous driving and mapping systems~\cite{ cui2024drive}. LMMs can revolutionize the conventional human-vehicle interaction paradigm~\cite{cui2024survey}. LMMs can process information from text and image inputs captured by in-vehicle cameras to understand complex traffic situations. In addition, they can significantly enhance personalized human-vehicle interactions through voice communication and user preference analysis. Drivers can use languages, gestures, and eyes to communicate their requests while driving, and LMMs provide real-time in-vehicle feedback via integrated visual displays. However, despite the unprecedented capabilities of LMMs, TSR-related studies based on LMMs remain unexplored. 

\section{Methodology}
In this section, we detail the proposed method for cross-domain zero-shot TSR, as illustrated in Fig.~\ref{fig2}. The proposed method begins with the localization and detection of traffic signs from original road images using a tailored extraction detector. Subsequently, we implement the proposed multi-step thinking strategy for stimulating the fine-grained TSR ability of LMMs.
\subsection{Traffic Sign Extraction}\label{sub3.1}
\subsubsection{\textbf{Segmentation}}\label{sub3.1.1}
\par In the proposed method,  the original road image $\mathcal{I}_o^i$ containing the traffic signs~$i \in \{0, 1,2, \ldots, N\}$ is segmented, where $N$ represents the number of traffic signs contained in the original road image. Specifically, the original road image $\mathcal{I}_o^i$ is input to a segmentation model, which generates segmentation images $\mathcal{I}_s^i$ with various object category labels for the original image. During traffic sign recognition, the traffic signs should be distinguished from other objects. Specifically, in the segmented image $\mathcal{I}_s^i$, each object category is coded as a different color for identification. We convert $\mathcal{I}_s^i$ to a mask image $\mathcal{I}_m^i$, thereby separating the traffic sign from the other objects and background in $\mathcal{I}_s^i$. The proposed method is unaffected by the architecture of the segmentation model, offering flexibility in implementation.
\subsubsection{\textbf{Extraction}}\label{sub3.1.2}
\par After segmentation, a custom extraction detector isolates the traffic signs. The extraction detector first obtains the coordinates of the traffic signs in the mask image $\mathcal{I}_m^i$ using a contour detection algorithm ~\cite{suzuki1985topological}. Then, the extraction detector uses the original road image $\mathcal{I}_o^i$ and the coordinates of the traffic signs to extract the image $\mathcal{I}_r^N$ that contains only the real traffic signs. $\mathcal{I}_r^N$ removes other objects and backgrounds in the original road image.
The extraction detector finally retrieves the traffic sign image $\mathcal{I}^i$ from $\mathcal{I}_r^N$ using the corresponding coordinates of the traffic signs. Here, $\mathcal{I}^i\in \mathbb{R}^{\mathcal{H} \times \mathcal{W} \times 3}$ represents the final extracted traffic sign image. Note that although $\mathcal{I}^i$ can also be obtained directly from the original road image $\mathcal{I}_o^i$ via the coordinates, the extracted traffic sign image contains unnecessary backgrounds. In contrast, the extraction detector removes backgrounds and avoids potential interference for subsequent recognition.
\begin{figure}[t]
        \centering
        \includegraphics[width=9cm]{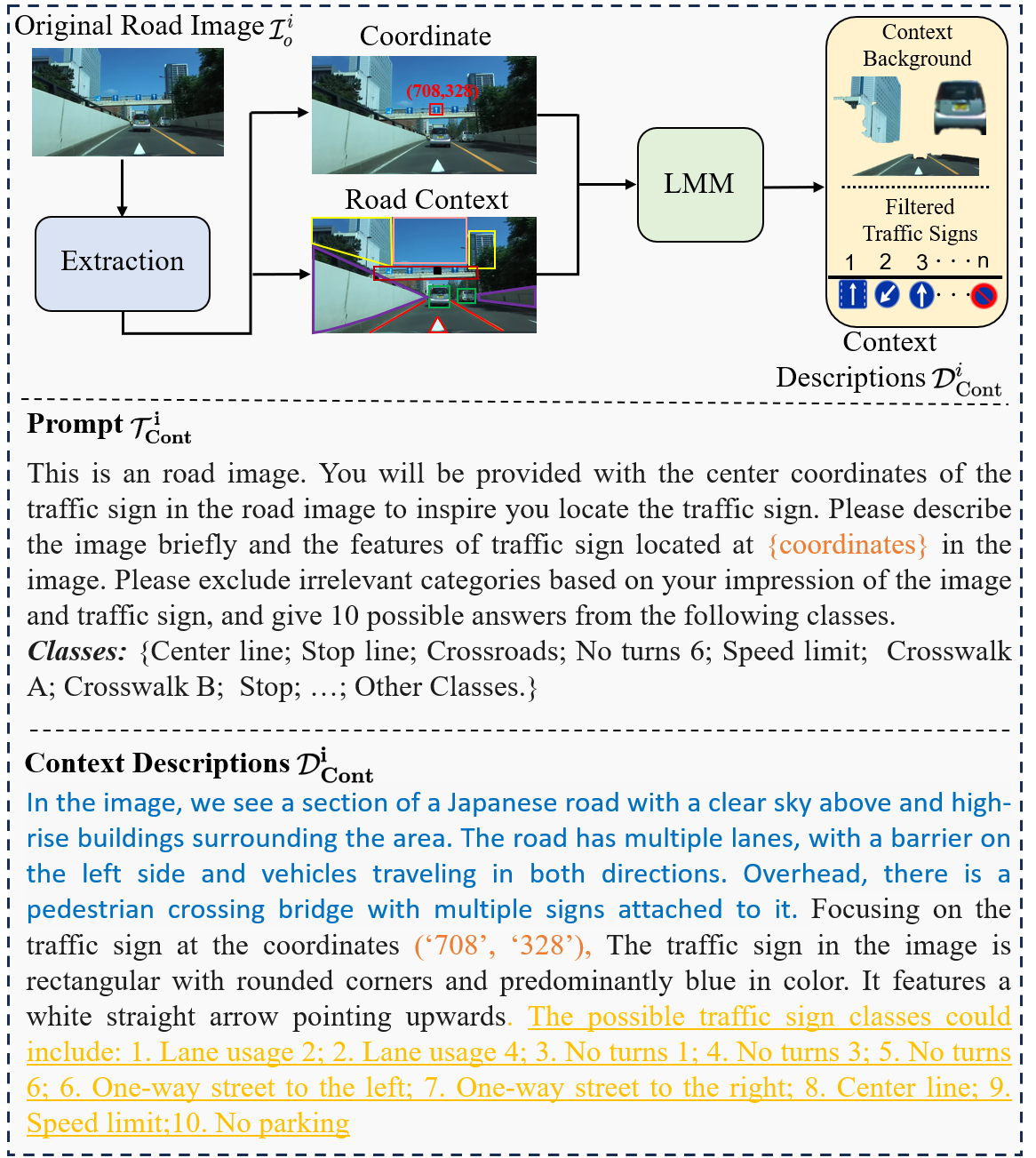}
        \caption{Generation of context descriptions. The extracted coordinates and road context of the target traffic sign help generate context descriptions, which include the coordinates (\textcolor{orange}{orange}) background and surrounding objects (\textcolor[RGB]{0,112,192}{blue}), and the prior traffic sign hypothesis (\textcolor[RGB]{255,192,0}{yellow}).}
        \label{fig3}
\end{figure}
\begin{figure}[t]
        \centering
        \includegraphics[width=9cm]{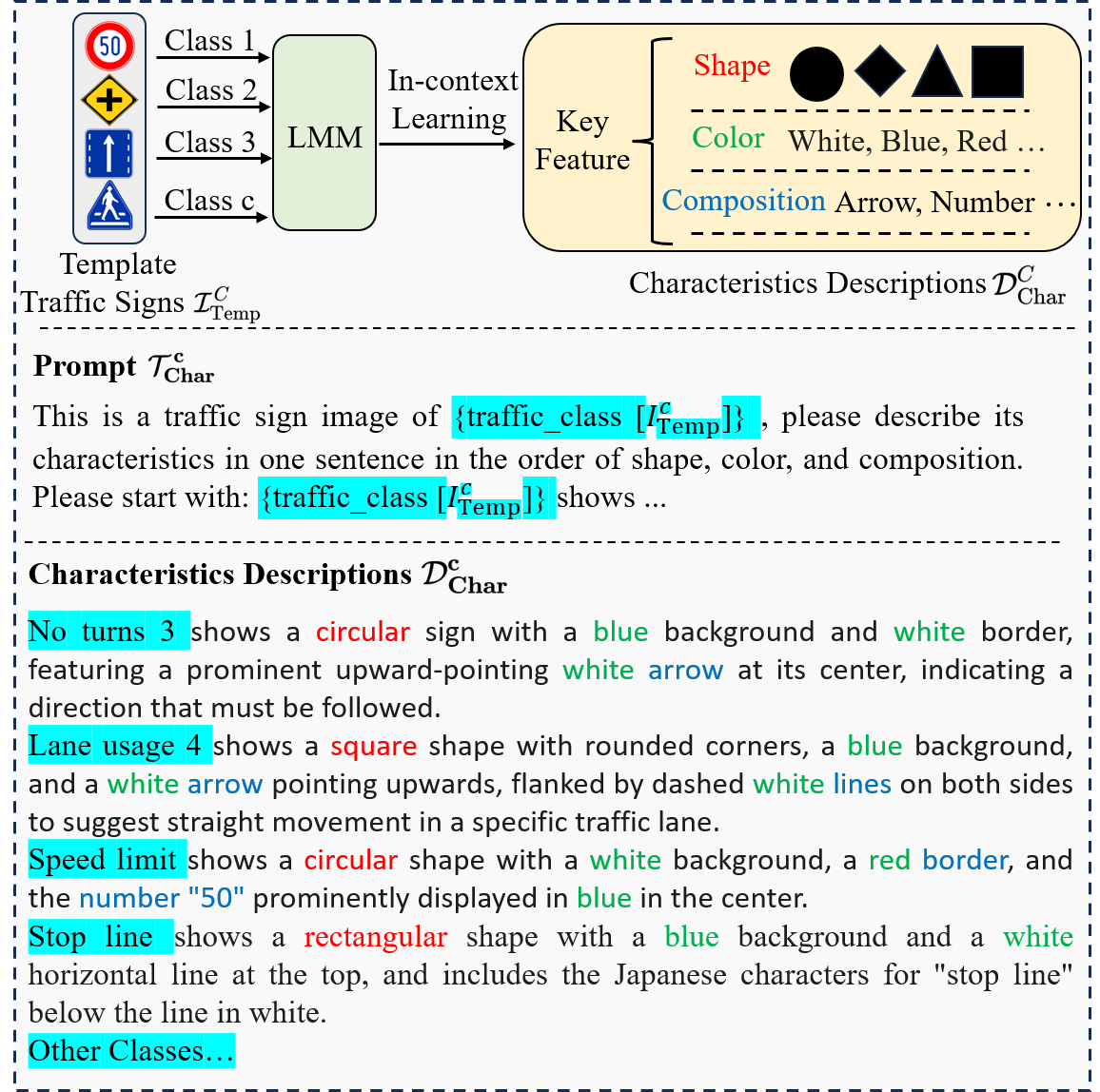}
        \caption{Generation of characteristic descriptions. We introduce in-context learning to help the LMM learn the key traffic sign features.}
        \label{fig4}
\end{figure}
\subsection{Milti-step Thinking}\label{sub3.2}
\par After obtaining the traffic sign image $\mathcal{I}^i$, we perform the multi-step thinking strategy to stimulate the perceptual potential of fine-grained TSR using the LMM. The proposed framework consists of two steps: prior knowledge generation and multi-step reasoning. 
\subsubsection{\textbf{Prior Knowledge Generation}}
\par In the proposed method, prior knowledge includes context descriptions of original road images, characteristic descriptions of template traffic signs, and differential descriptions of similar traffic signs.
The inputs of LMMs are typically an image $\mathcal{I}^i$ and a text query $\mathcal{T}^i = [t^i_1, \ldots, t^i_{l_i}]$ with length $l_i$, and LMMs generate a sequence of textual output $\mathcal{T}_\text{out}^i = [t_1^i, \ldots, t_{l_o}^i]$ with length $l_o$ as follows:
\begin{equation}
\mathcal{T}_\text{out}^i = \mathrm{LMM}(\mathcal{I}^i, \mathcal{T}^i). 
\end{equation}
\newline\textbf{Context Descriptions}: Original road images contain important contextual information about traffic signs; thus, we transform these images into context descriptions to fully use the scene information. Given an original road image $\mathcal{I}_o^i$, the context descriptions $\mathcal{D}^i_\text{Cont}=[\mathcal{D}^{i}_\text{Cont}, ...,\mathcal{D}^{N}_\text{Cont}]$ are generated as follows:
\begin{equation}
\mathcal{D}^i_\text{Cont} = \mathrm{LMM}({\mathcal{I}_o^i},~\mathcal{T}^i_\text{Cont}),
\end{equation}
where $\mathcal{T}^i_\text{Cont}$ represents the prompt for generating the context descriptions. As shown in Fig.~\ref{fig3}, we carefully designed $\mathcal{T}^i_\text{Cont}$ so that the generated contextual descriptions contain the context background information understood by the LMM from the original road image. In addition, as in the real-world question-answering process, we find that narrowing the range of answers can reduce the recognition difficulty of the LMM. Therefore, we propose a prior traffic sign hypothesis, which allows the LMM to filter irrelevant traffic sign types and provide potential candidates. 
Similar to human cognition, where irrelevant answers are swiftly filtered based on existing knowledge, the potential traffic sign candidates generated by the prior traffic sign hypothesis are obtained from the preliminary understanding of the original road image of the LMM. This preliminary understanding stimulates subsequent detailed thinking. In addition, when multiple traffic signs exist in the original road image, it is difficult for the LMM to perform context description and prior traffic sign hypothesis generation. Therefore, we simplify the complex process and propose a prompt optimization method based on center coordinates. The proposed prompt optimization method provides the center coordinates of traffic signs to inspire the LMM to locate the target traffic sign from the original road image. The center coordinates are obtained from the extraction detector; thus, no additional calculations for center coordinates are required. The center coordinates help the LMM locate the target traffic sign and generate corresponding background descriptions and prior traffic sign hypotheses.
\newline\textbf{Characteristic Descriptions}: Fine-grained TSR poses a challenge for LMMs, because their existing knowledge typically struggles to accurately identify specific traffic sign types. Leveraging the accessibility of template traffic signs from national databases, we reduce reliance on extensive training data. Unlike previous methods that match templates at the feature level, where real-world signs vary due to lighting, angles, and occlusions, thereby increasing cross-domain gaps, we introduce in-context learning to generate characteristic descriptions \(\mathcal{D}_\text{Char} = [\mathcal{D}^1_\text{Char}, \ldots, \mathcal{D}^C_\text{Char}]\) for each class \(c\) of template traffic signs \(\mathcal{I}_\text{Temp} = [\mathcal{I}^1_\text{Temp}, \ldots, \mathcal{I}^C_\text{Temp}]\). This is achieved with prompts \(\mathcal{T}_\text{Char} = [\mathcal{T}^1_\text{Char}, \ldots, \mathcal{T}^C_\text{Char}]\) as follows:
\begin{equation}
\mathcal{D}^c_\text{Char} = \mathrm{LMM}(\mathcal{I}^c_\text{Temp}, \mathcal{T}^c_\text{Char}), \label{eq:char}
\end{equation}
where \(\mathcal{T}^c_\text{Char}\) denotes the prompt tailored for \(\mathcal{I}^c_\text{Temp}\).

As shown in Fig.~\ref{fig4}, traffic signs universally exhibit three key attributes: shape, color, and composition. Our prompts (see Fig.~\ref{fig4}) guide the LMM to focus on these features, avoiding extraneous details. This in-context learning generates each \(\mathcal{D}^c_\text{Char}\) only once and stores them in a memory bank. By circumventing feature-level computation, our approach mitigates cross-domain disparities between templates and real-world signs. The prompts are simple and uniform and require no class-specific tuning, thereby enhancing efficiency and scalability.
\begin{figure}[t]
        \centering
        \includegraphics[width=8.5cm]{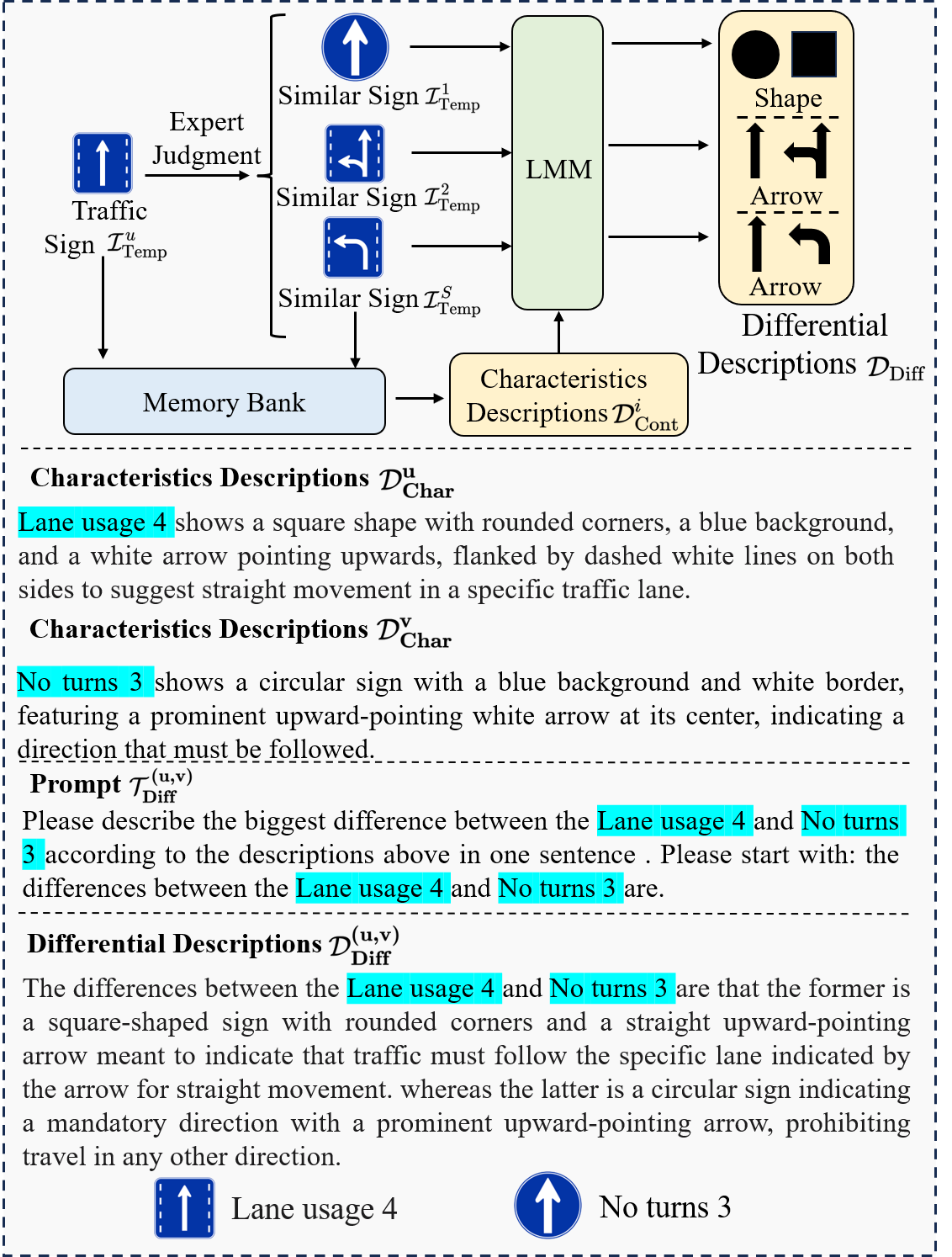}
        \caption{Differential description generation. Differences between similar traffic signs are emphasized to strengthen the fine-grained thinking ability of the LMM.}
        \label{fig5}
\end{figure}
\newline\textbf{Differential Descriptions}: Certain traffic signs share highly similar characteristics, complicating fine-grained recognition. To address this issue, differential descriptions are generated to highlight subtle distinctions. For a template sign \(\mathcal{I}^u_\text{Temp}\) and a similar sign \(\mathcal{I}^v_\text{Temp} \in [\mathcal{I}^1_\text{Temp}, \ldots, \mathcal{I}^S_\text{Temp}]\), we first obtain their characteristic descriptions using Eq.~\eqref{eq:char}:
\begin{equation}
\mathcal{D}^u_\text{Char} = \mathrm{LMM}(\mathcal{I}^u_\text{Temp}, \mathcal{T}^u_\text{Char}), \label{eq:char_u}
\end{equation}
\begin{equation}
\mathcal{D}^v_\text{Char} = \mathrm{LMM}(\mathcal{I}^v_\text{Temp}, \mathcal{T}^v_\text{Char}). \label{eq:char_v}
\end{equation}
The differential description \(\mathcal{D}^{u,v}_\text{Diff}\) is then derived as follows:
\begin{equation}
\mathcal{D}^{u,v}_\text{Diff} = \mathrm{LMM}(\mathcal{D}^u_\text{Char}, \mathcal{D}^v_\text{Char}, \mathcal{T}^{u,v}_\text{Diff}), \label{eq:diff}
\end{equation}
where \(\mathcal{T}^{u,v}_\text{Diff}\) denotes the prompt designed to elicit differences. The complete set of differential descriptions is given by:
\begin{equation}
\mathcal{D}_\text{Diff} = \bigcup_{u, v \in \{1, \ldots, S\}} \mathcal{D}^{u,v}_\text{Diff}. \label{eq:diff_set}
\end{equation}
As shown in Fig.~\ref{fig5}, experts identify similar sign pairs, and the characteristic descriptions in the memory bank inform the generation of \(\mathcal{D}_\text{Diff}\). These descriptions emphasize nuanced differences (e.g., lane usage vs. no turns), refining the fine-grained recognition capabilities of the LMM.

\begin{algorithm}[t]
\caption{Cross-domain Multi-step Thinking (CdMT) for TSR}
\label{alg:cdmt}
\begin{algorithmic}[1]
\Require Raw road image \(\mathcal{I}_o^i\), template signs \(\mathcal{I}_\text{Temp}\), prompts \(\mathcal{T}_\text{Cont}\), \(\mathcal{T}_\text{Char}\), \(\mathcal{T}_\text{Diff}\), \(\mathcal{T}_\text{Multi}\)
\Ensure Recognized traffic sign type \(\mathcal{T}^i_\text{out}\)
\Statex \textbf{Phase 1: Traffic Sign Extraction}
\State \(\mathcal{I}_s^i \gets \text{Segment}(\mathcal{I}_o^i)\) \Comment{Segment raw image}
\State \(\mathcal{I}_m^i \gets \text{ConvertToMask}(\mathcal{I}_s^i)\) \Comment{Generate mask}
\State \(\text{Coords} \gets \text{ContourDetect}(\mathcal{I}_m^i)\) \Comment{Extract coordinates}
\State \(\mathcal{I}_r^N \gets \text{Extract}(\mathcal{I}_o^i, \text{Coords})\) \Comment{Refine image}
\State \(\mathcal{I}^i \gets \text{Retrieve}(\mathcal{I}_r^N, \text{Coords})\) \Comment{Get target sign}
\Statex \textbf{Phase 2: Prior Knowledge Generation}
\State \(\mathcal{D}^i_\text{Cont} \gets \mathrm{LMM}(\mathcal{I}_o^i, \mathcal{T}_\text{Cont} + \text{Coords})\) \Comment{Context with coords}
\For{\(c = 1\) to \(C\)} \Comment{For each template class}
    \State \(\mathcal{D}^c_\text{Char} \gets \mathrm{LMM}(\mathcal{I}^c_\text{Temp}, \mathcal{T}^c_\text{Char})\)
    \State Store \(\mathcal{D}^c_\text{Char}\) in memory bank
\EndFor
\For{each similar pair \((u, v)\) in \(\mathcal{I}_\text{Temp}\)} \Comment{Expert-identified}
    \State \(\mathcal{D}^{u,v}_\text{Diff} \gets \mathrm{LMM}(\mathcal{D}^u_\text{Char}, \mathcal{D}^v_\text{Char}, \mathcal{T}^{u,v}_\text{Diff})\)
    \State \(\mathcal{D}_\text{Diff} \gets \mathcal{D}_\text{Diff} \cup \mathcal{D}^{u,v}_\text{Diff}\)
\EndFor
\Statex \textbf{Phase 3: Multi-step Reasoning}
\State \(\mathcal{T}^i_\text{out} \gets \mathrm{LMM}(\mathcal{I}^i, \mathcal{D}^i_\text{Cont}, \mathcal{D}_\text{Char}, \mathcal{D}_\text{Diff}, \mathcal{T}_\text{Multi})\)
\State \Return \(\mathcal{T}^i_\text{out}\)
\end{algorithmic}
\end{algorithm}

\begin{table*}[ht]
\centering
\small
\setlength{\tabcolsep}{2.7pt} % 调整列间距
\caption{Top-$k$ zero-shot fine-grained TSR performance on five datasets. We compare the proposed method with state-of-the-art methods. Bold represents the best result, and an underline represents the second-best result. Note that the presented results are the average accuracy obtained over five trials.}
\begin{tabular}{l|ccc|ccc|ccc|ccc|ccc}
\toprule

\multirow{2}{*}{\textbf{Method}}
&\multicolumn{3}{c|}{\textbf{GTSRB}} 
& \multicolumn{3}{c|}{\textbf{BTSD}} 
& \multicolumn{3}{c|}{\textbf{TT-100K}}
& \multicolumn{3}{c|}{\textbf{Sapporo}} 
& \multicolumn{3}{c}{\textbf{Yokohama}}\\
\cline{2-16}
% \midrule
& \textbf{Top-1} & \textbf{Top-3} & \textbf{Top-5} & \textbf{Top-1} & \textbf{Top-3} & \textbf{Top-5} & \textbf{Top-1} & \textbf{Top-3} & \textbf{Top-5} & \textbf{Top-1} & \textbf{Top-3} & \textbf{Top-5} &\textbf{Top-1} & \textbf{Top3} & \textbf{Top-5}\\
\midrule
Song et al.~\cite{yucong2021traffic} 
            &0.10     &0.23     &0.29 
            & 0.19  	&0.25     &0.32 
            &0.04   	&0.10     &0.13 
            &0.04   	&0.57    &0.77 
            &0.04   	&0.18     &0.42 
\\
Ren et al.~\cite{ren2009general}
            &0.41     &0.64     &0.77 
            &0.11   	&0.36     &0.50 
            & 0.26  	&0.42     &0.50 
            &0.34   &0.47     &0.50 
            &0.21   	&0.42     &0.48 
\\

% VGG
%             & 0.07    &0.21     &0.26 
%             &0.05   	&0.11     &0.21 
%             &   	&     & 
%             &0.02   &0.10     &0.18 
%             &0.01   	&0.16     &0.31
% \\

Gan et al.~\cite{gan2023zero}   
            &0.56     &0.76     &0.84 
            &0.67     &0.84     &0.91 
            &0.12     &0.22     &0.36 
            & 0.42    & 0.71    &0.79 
            &0.27     &0.42     &0.51 
\\
% Hao et al. Inception-V4~\cite{haloi2015traffic}
%             &     &     & 
%             &   	&     & 
%             &   	&     & 
%             &   	&     &
%         &   	&     & 
%  \\
DenseNet-121~\cite{huang2017densely}  
            &0.31     &0.46     &0.59 
            &0.21   	&0.32     &0.49 
            &0.08   	&0.14     &0.24 
            &0.73   	&0.82     &0.84 
            &0.23   	&0.47     &0.70
\\

EfficientNet-B0~\cite{tan2019efficientnet}
            &0.52     &0.76     &0.90 
            &0.60   	&0.86     &\underline{0.93} 
            & 0.17  	&0.30     &0.38 
            &0.51   	&0.66     &0.74 
            &0.25   	&0.44     &0.60
\\

%Mobilenet-V3
Li et al.~\cite{li2018real}
            & 0.75  	&0.83     &0.89 
            &0.82   	&\textbf{0.91}     &\textbf{0.94} 
            &0.27   	& 0.46    &0.60 
            &0.70   	&0.80     &0.83 
            & 0.29  	&0.45     &0.69
\\            
% Sill-Net~\cite{howard2019searching}*
%             &     &     & 
%             &   	&     & 
%             &   	&     & 
%               &   	&     & 
%             &   	&     &
% \\ 
% Zheng et al. (ViT-B)~\cite{zheng2022evaluation}* 
%             &0.32     & 0.45    &0.53 
%             &0.11   	&0.20     &0.35 
%             & 0.14  	&0.31     &0.38 
%             &0.22   	&0.58     &0.74 
%             &0.12   	&0.34     &0.47
% \\
Zheng et al. (ViT-L)~\cite{zheng2022evaluation} 
            &0.44     &0.58     &0.70 
            &0.39   	&0.57     &0.64 
            &0.09   	&0.16     &0.21 
            &0.54   	& 0.63    &0.75 
            &0.19  	&0.36     &0.44
\\
Luo et al.~\cite{luo2023pre} 
& 0.15    & 0.35    &0.48 
&0.22   	&0.27     &0.34 
&0.14   	&0.28     & 0.41
& 0.39 	&0.57     &0.70 
&  0.18 	&0.35    &0.56
\\
MobileViT~\cite{mehta2021mobilevit}
&0.05     &0.11     &0.22 
& 0.02  	&0.07     & 0.10
&0.05   	&0.11     &0.15 
&0.08  	&0.10     &0.29 
            & 0.06  	&0.35     &0.42
\\
%Swin-Transformer

Swin-Transformer V2~\cite{liu2022swin}  
&0.14     &0.26     &0.37 
& 0.06  	&0.17     &0.32 
& 0.09  	&0.17     &0.23 
&0.06  	&0.10     &0.18 
            &0.09   	&0.27     &0.58
\\
MAE~\cite{he2022masked}
            &0.20     &0.32     &0.47 
            &0.13   	&0.36     &0.49 
            & 0.06  	&0.10     &0.13 
            &0.14   	&0.27     &0.41
            &0.17   	&0.32     &0.51
\\
DeiT~\cite{touvron2021training}
&0.27     & 0.45    &0.57 
&0.12   	&0.28     &0.42 
&0.34   	&0.60     &0.70 
&0.71  	&0.83     &0.88 
            &0.26   	&0.47     &0.69
\\
% CLIP (ResNet-50)~\cite{radford2021learning}
%             &     &     & 
%             &   	&     & 
%             &   	&     & 
%             &0.27   	&0.50     &0.57 
 
%             &   	&     &
% \\
CLIP (ViT-B/32)~\cite{radford2021learning}
            &0.24     & 0.35    &0.48 
            & 0.20  	&0.30     &0.38 
            &0.29   	&0.50     &0.62 
            &0.27   	&0.50     &0.57 
             & 0.14  	&0.48     &0.60 
\\
% \textcolor{black}{CoOp~\cite{zhou2022learning}
%             &0.32     & 0.44    &0.63 
%             & 0.20  	&0.30     &0.38 
%             &0.29   	&0.50     &0.62 
%             &0.27   	&0.50     &0.57 
%             & 0.14  	&0.48     &0.60}  
% \\
% \textcolor{black}{MaPLe~\cite{khattak2023maple}
%             &0.24     & 0.35    &0.48 
%             & 0.20  	&0.30     &0.38 
%             &0.29   	&0.50     &0.62 
%             &0.27   	&0.50     &0.57 
%              & 0.14  	&0.48     &0.60 }
% \\
% \textcolor{black}{CLIP-Adapter~\cite{gao2024clip}
%             &0.24     & 0.35    &0.48 
%             & 0.20  	&0.30     &0.38 
%             &0.29   	&0.50     &0.62 
%             &0.27   	&0.50     &0.57 
%              & 0.14  	&0.48     &0.60 }
% \\

\textcolor{black}{CoOp~\cite{zhou2022learning}}
  & \textcolor{black}{0.32} & \textcolor{black}{0.44} & \textcolor{black}{0.63} 
  & \textcolor{black}{0.25} & \textcolor{black}{0.36} & \textcolor{black}{0.55} 
  & \textcolor{black}{0.36} & \textcolor{black}{0.58} & \textcolor{black}{0.71} 
  & \textcolor{black}{0.33} & \textcolor{black}{0.62} & \textcolor{black}{0.74} 
  & \textcolor{black}{0.17} & \textcolor{black}{0.56} & \textcolor{black}{0.65}
\\
\textcolor{black}{MaPLe~\cite{khattak2023maple}}
  & \textcolor{black}{0.28} & \textcolor{black}{0.35} & \textcolor{black}{0.49} 
  & \textcolor{black}{0.23} & \textcolor{black}{0.32} & \textcolor{black}{0.41} 
  & \textcolor{black}{0.37} & \textcolor{black}{0.62} & \textcolor{black}{0.76} 
  & \textcolor{black}{0.34} & \textcolor{black}{0.66} & \textcolor{black}{0.79} 
  & \textcolor{black}{0.20} & \textcolor{black}{0.61} & \textcolor{black}{0.72}
\\
\textcolor{black}{CLIP-Adapter~\cite{gao2024clip}}
  & \textcolor{black}{0.37} & \textcolor{black}{0.52} & \textcolor{black}{0.71}
  & \textcolor{black}{0.32} & \textcolor{black}{0.43} & \textcolor{black}{0.61}
  & \textcolor{black}{0.43} & \textcolor{black}{0.69} & \textcolor{black}{0.83}
  & \textcolor{black}{0.41} & \textcolor{black}{0.69} & \textcolor{black}{0.83}
  & \textcolor{black}{0.26} & \textcolor{black}{0.63} & \textcolor{black}{0.75}
\\

EVA-02~\cite{fang2023eva}
&0.41     &0.67     &0.75 
&0.30   	&0.51     &0.66 
&0.32   	&0.61     &0.76 
&0.48  	&0.53     &0.62 
            &0.29   	&0.46     &0.70
\\
LLaVA-1.5~\cite{liu2023visual}
            &0.32     &0.45     &0.48 
            &0.28   	&0.33     &0.42 
            &0.13   	&0.21     &0.38 
            &0.09   	&0.20     &0.46 
            &0.11   	&0.32     &0.43 

\\
LLaVA-NeXT~\cite{liu2024llavanext}
            &0.39     &0.48     &0.57 
            &0.31   	&0.38     &0.46 
            &0.20   	&0.31     &0.52 
            &0.10   	&0.23     &0.51 
            &0.13   	&0.38     &0.47 
\\
VITA-1.5~\cite{fu2025vita}
            &0.45     &0.56     &0.63 
            &0.39   	&0.48     &0.56 
            &0.25   	&0.42     &0.63 
            &0.18   	&0.31     &0.40 
            &0.20   	&0.45     &0.72 
\\
Gpt-4v~\cite{openaisystem}
            &0.81     &0.85     &0.87 
            & 0.70  	&0.83     &0.87 
            &0.72   	&0.82     &0.86 
            &0.32   	&0.39     &0.47 
            & 0.22  	&0.62     & 0.68
\\

Gpt-4o~\cite{openaisystem1}
            &0.89   	&0.89     &0.90 
            &0.83   	&0.86     &0.87 
            &0.74   	&0.83     &0.86 
            &0.57   	&0.69     &0.78 
            &0.49   	&0.71     &0.83 
\\

\textbf{CdMT-LLaVA-1.5}
            &0.55   &0.71     &0.83 
            &0.48   	&0.60     &0.73 
            &0.45   	&0.61     &0.76 
            &0.35   	&0.46     &0.74 
            &0.36   	&0.52     &0.81             
\\
\textbf{CdMT-LLaVA-NeXT}
            &0.60     &0.74     &0.85 
            &0.51   	&0.65     &0.80 
            &0.48   	&0.62     &0.77 
            &0.37   	&0.50     &0.77 
            &0.41   	&0.55     &0.84 
\\
\textbf{CdMT-VITA-1.5}
            &0.68     &0.83     &0.90 
            &0.65   	&0.76     &0.82 
            &0.52   	&0.67     &0.80 
            &0.45   	&0.63     &0.82 
            &0.61   	&0.75     &0.87 
\\
\textbf{CdMT-Gpt-4v}
            &\underline{0.91}  & \underline{0.96}  & \underline{0.97}
            &\textbf{0.89}  &  \textbf{0.91} & 0.92  
            & \underline{0.90}  & \underline{0.97}  & \textbf{0.99}
            &\underline{0.77}  & \underline{0.86}  & \underline{0.89}
            & \underline{0.83}  & \underline{0.91}  & \underline{0.95}

            %             &\textbf{0.91}  & \textbf{0.96}  & \textbf{0.97}
            % &\textbf{0.89}  &  \textbf{0.91} & 0.92  
            % &\textbf{0.90}  & \textbf{0.97}  & \textbf{0.99}
            % &\textbf{0.77}  &\textbf{0.86}   &\textbf{0.89} 
            % &\textbf{0.83}  &\textbf{0.91}   &\textbf{0.95} 
  \\          
\textbf{CdMT-Gpt-4o}
            &\textbf{0.93}  & \textbf{0.97}  & \textbf{0.98}
            &\underline{0.88}  & \textbf{0.91} & 0.91  
            &\textbf{0.97}  & \textbf{0.99}  & \textbf{0.99}
            & \textbf{0.89}  & \textbf{0.95}   &\textbf{1.00} 
            &\textbf{0.85}  &\textbf{0.96}   &\textbf{0.97} 
\\
% \textbf{Ours (o1-mini)}
%                         & -  	& -    &- 
%             &\textbf{0.90}  & \textbf{0.92} & \textbf{0.94}  
%                         & -  	&-     &- 
%             &\textbf{0.91}  &\textbf{0.95}   &\textbf{1.00} 
%                         &-   	&-     &- 
% \\
\bottomrule
\end{tabular}
\label{tab2}
\end{table*}

\begin{figure*}[ht]{
    \centering
    \includegraphics[width=0.9\textwidth]{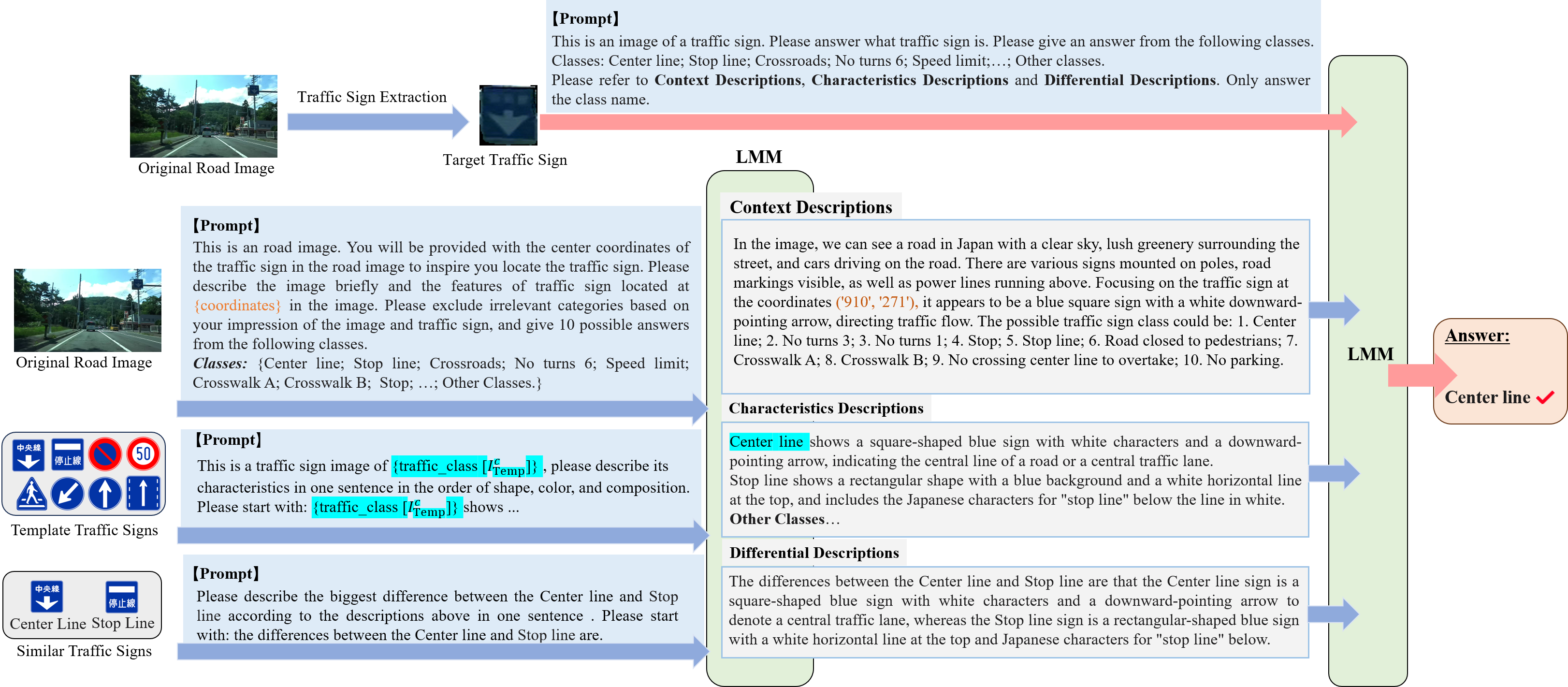}
    \caption{Recognition examples of the proposed CdMT framework on the Sapporo dataset.} \label{fig6}}
\end{figure*}

\begin{figure*}[ht]{
    \centering
    \includegraphics[width=0.9\textwidth]{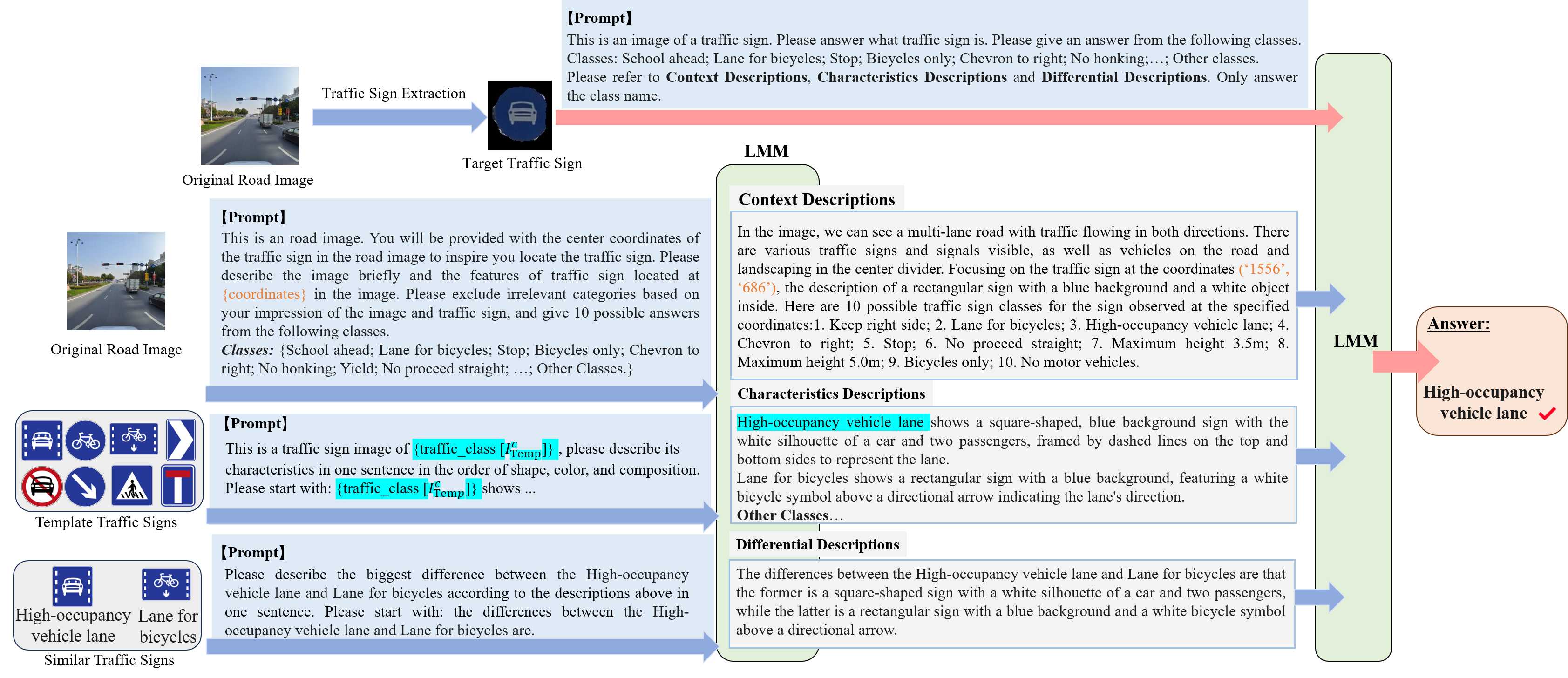}
    \caption{Recognition examples of the proposed CdMT framework on the TT-100K dataset.} \label{supb5}}
\end{figure*}

\begin{table*}[ht]
\centering
\small

\setlength{\tabcolsep}{1.5pt} % 调整列间距
\renewcommand{\arraystretch}{1.1} % 调整行间距
\caption{Top-$k$ zero-shot fine-grained TSR performance based on different thinking strategies. ``Cont*,'' ``Char*,'' and ``Diff*'' represent the context, characteristic, and differential descriptions, respectively. Bold represents the best result. ``-'' indicates that no context descriptions are generated due to the lack of original road images.}
\begin{tabular}{c|c|c|c|ccc|ccc|ccc|ccc|ccc}
\toprule

\multirow{2}{*}{\textbf{LMM}}& \multirow{2}{*}{\textbf{Cont*}}&\multirow{2}{*}{\textbf{Char*}}&\multirow{2}{*}{\textbf{Diff*}}&\multicolumn{3}{c|}{\textbf{GTSRB}} 
& \multicolumn{3}{c|}{\textbf{BTSD}} 
& \multicolumn{3}{c|}{\textbf{TT-100K}}
& \multicolumn{3}{c|}{\textbf{Sapporo}} 
& \multicolumn{3}{c}{\textbf{Yokohama}}\\
%\midrule
\cline{5-19}
&& & & \textbf{Top-1} & \textbf{Top-3} & \textbf{Top-5} & \textbf{Top-1} & \textbf{Top-3} & \textbf{Top-5} & \textbf{Top-1} & \textbf{Top-3} & \textbf{Top-5} & \textbf{Top1} & \textbf{Top-3} & \textbf{Top-5} &\textbf{Top-1} & \textbf{Top-3} & \textbf{Top-5}\\
\midrule
\multirow{1}{*}{\textbf{Gpt-4v}}&& &
            &0.81     &0.85     &0.87 
            & 0.70  	&0.83     &0.87 
            &0.72   	&0.82     &0.86 
            &0.32   	&0.39     &0.47 
            & 0.22  	&0.62     & 0.68
            \\

\cline{0-0}
\multirow{7}{*}{\textbf{CdMT-Gpt-4v}} &
\checkmark & & 
            &-     &-     &- 
            &-     &-     &- 
            &0.77   	& 0.86    &0.88 
            &0.48   	&0.60     & 0.68
            &0.49   	&0.78     & 0.91
\\
& & \checkmark &
            &0.87     &0.95     &0.96 
            &0.87   &0.90     &0.91 
            &0.84   	&0.90     & 0.91
            &0.55   	&0.65     &0.74 
            &0.66   	&0.74     &0.79 
 \\

&&&\checkmark
            &0.82     &0.87     &0.88 
            &0.76   	&0.86     &0.88 
            &0.77   	&0.85     & 0.88
            &0.42  &0.54     &0.66 
            &0.35   	&0.64     &0.77 
 \\
&\checkmark&&\checkmark 
            &-     &-     &- 
            &-     &-     &- 
            &0.76   	&0.85     &0.89 
            &0.62   	&0.74     &0.78 
            & 0.55  	&0.83     &0.91 
\\
&\checkmark&\checkmark &  
            &-     &-     &- 
            &-     &-     &- 
            &0.85   	&0.92     &0.92 
            &0.76   &0.84     &0.86 
            &0.66   	&0.87     &0.94 
\\

&& \checkmark & \checkmark
            &\textbf{0.91}     &\textbf{0.96}     &\textbf{0.97} 
            &\textbf{0.89}   	&\textbf{0.91}     &\textbf{0.92} 
            &0.88   	&0.94     & 0.95
            &0.68   	&0.82     &0.87 
            &0.81   	&0.88     &0.94 
\\

&\checkmark & \checkmark & \checkmark
            &-     &-     &- 
            &-     &-     &- 
            &\textbf{0.90}  	&\textbf{0.97}     &\textbf{0.99}
            &\textbf{0.77} 	&\textbf{0.86} &\textbf{0.89}  
            &\textbf{0.83}  &\textbf{0.91} &\textbf{0.95} 
\\
\hline
\hline
\multirow{1}{*}{\textbf{Gpt-4o}}&& &
            &0.89   	&0.89     &0.90 
            &0.83   	&0.86     &0.87 
            &0.74   	&0.83     &0.86 
            &0.57   	&0.69     &0.78 
            &0.49   	&0.71     &0.83 
            
\\
\cline{0-0}
\multirow{7}{*}{\textbf{CdMT-Gpt-4o}}
&\checkmark & & 
            &-     &-     &- 
            &-     &-     &- 
            &0.82   	&0.91     &0.93 
            &0.77   	&0.79     &0.83 
            &0.50   	&0.83     &0.89 
\\
& & \checkmark &
            &0.92     &0.96     &0.98 
            &0.86  	&0.88     &0.88 
            &0.93   	&0.98     &0.98 
            &0.86   	&0.91     &0.95 
            &0.82   	&0.93     &\textbf{0.97} 
 \\

&&&\checkmark
            &0.89     &0.95     &0.95 
            &0.85   	&0.89     &0.89 
            &0.92   	&0.97     &0.97 
            &0.74   	&0.85     &0.92 
            &0.58   	&0.74     &0.85 
 \\
&\checkmark&&\checkmark 
            &-     &-     &- 
            &-     &-     &- 
            &0.93     &0.97     &0.97 
            &0.85   	&0.91     &0.93 
            &0.68   	&0.85     &0.90
\\
&\checkmark&\checkmark &  
            &-     &-     &- 
            &-     &-     &- 
            &0.95     &0.98     &0.98 
            &0.87   	&0.93     &0.96 
            &0.79   	&0.94     &0.96
\\

&& \checkmark & \checkmark
            &\textbf{0.93}     &\textbf{0.97}  &\textbf{0.98} 
            &\textbf{0.88}   	&\textbf{0.91}     &\textbf{0.91} 
            &0.96   	&0.98     &\textbf{0.99}
            &\textbf{0.89}   	&0.94     &0.99 
            &0.82   	&0.94     &0.96 
\\

&\checkmark & \checkmark & \checkmark
            &-     &-     &- 
            &-     &-     &- 
            &\textbf{0.97}     &\textbf{0.99}     &\textbf{0.99} 
            &\textbf{0.89}   	&\textbf{0.95}     &\textbf{1.00} 
            &\textbf{0.85}   	&\textbf{0.96}     &\textbf{0.97}
\\

\bottomrule
\end{tabular}
\label{tab3}
\end{table*}

\begin{table*}[ht]
\centering
\small
\setlength{\tabcolsep}{4pt} % 调整列间距
\renewcommand{\arraystretch}{1} % 调整行间距
\caption{Top-$k$ zero-shot fine-grained TSR performance based on different context description generation methods. Bold represents the best result.}
\begin{tabular}{c|c|c|ccc|ccc|ccc}
\toprule

\multirow{2}{*}{\textbf{LMM}}&\multirow{2}{*}{\textbf{Prior hypothesis}}&\multirow{2}{*}{\textbf{Center coordinates}}
& \multicolumn{3}{c|}{\textbf{TT-100K}}
& \multicolumn{3}{c|}{\textbf{Sapporo}} 
& \multicolumn{3}{c}{\textbf{Yokohama}}\\
%\midrule
\cline{4-12}
&& & \textbf{Top-1} & \textbf{Top-3} & \textbf{Top-5} & \textbf{Top-1} & \textbf{Top-3} & \textbf{Top-5} &\textbf{Top-1} & \textbf{Top-3} & \textbf{Top-5}\\
\midrule
\multirow{4}{*}{\textbf{CdMT-Gpt-4v}}
&& 
            &0.87     &0.92     &0.92 
            &0.68   	&0.86     &0.88 
            &0.78   	&0.84     &0.88 
\\
&\checkmark &
            &0.86     &0.92     &0.93 
            &0.60   	&0.76     &0.76
            &0.73   	&0.88     &0.91 
\\
&& \checkmark 
            &0.85     &0.93     &0.95 
            & 0.67  	&\textbf{0.87}     &0.88 
            &0.74   	&0.88     &0.91 
 
 \\

&\checkmark&\checkmark 
            &\textbf{0.90}  	&\textbf{0.97}     &\textbf{0.99}
            &\textbf{0.77} 	&  0.86 &\textbf{0.89}  
            &\textbf{0.83}   	&\textbf{0.91}     &\textbf{0.95} 
 \\
\hline

\multirow{4}{*}{\textbf{CdMT-Gpt-4o}}
&& 
            &0.90     &0.95     &0.98 
            &0.80   	&0.88     &0.89 
            &0.80   	&0.92     &0.95 
\\
&\checkmark &
            &0.86     &0.90     &0.93 
            &0.75   	&0.82     &0.84 
            &0.77   	&0.90     &0.92 
\\
&& \checkmark 
            &0.90     &0.95     &0.97 
            &0.78   	&0.88     &0.90 
            &0.79   	&0.92     &0.95 
 
 \\

&\checkmark&\checkmark 
            &\textbf{0.97}     &\textbf{0.99}     &\textbf{0.99} 
            &\textbf{0.89}   	&\textbf{0.95}     &\textbf{1.00} 
            &\textbf{0.85}   	&\textbf{0.96}     &\textbf{0.97} 
            
\\
\bottomrule
\end{tabular}
\label{tab4}
\end{table*}

\begin{table*}[htbp]
\centering
\small

\setlength{\tabcolsep}{1.5pt} % 调整列间距
\renewcommand{\arraystretch}{1.2} % 调整行间距
\caption{TSR results based on different thinking steps. ``w'' and ``w/o'' represent the cases in which the thinking steps are changed and are not changed, respectively.}
\begin{tabular}{c|c|ccc|ccc|ccc|ccc|ccc}
\toprule
\multirow{2}{*}{\textbf{LMM}}&\multirow{2}{*}{\textbf{Change Thinking}}&\multicolumn{3}{c|}{\textbf{GTSRB}} 
& \multicolumn{3}{c|}{\textbf{BTSD}} 
& \multicolumn{3}{c|}{\textbf{TT-100K}}
& \multicolumn{3}{c|}{\textbf{Sapporo}} 
& \multicolumn{3}{c}{\textbf{Yokohama}}
\\
%\midrule
\cline{3-17}
 &&\textbf{Top-1} & \textbf{Top-3} & \textbf{Top-5} & \textbf{Top-1} & \textbf{Top-3} & \textbf{Top-5} & \textbf{Top-1} & \textbf{Top-3} & \textbf{Top-5} & \textbf{Top-1} & \textbf{Top-3} & \textbf{Top-5} &\textbf{Top-1} & \textbf{Top-3} & \textbf{Top-5}\\
\midrule
\multirow{2}{*}{\textbf{CdMT-Gpt-4v}}
            &w
            &0.91  & 0.96  & 0.97
            &0.89  & 0.91  & 0.92  
            &0.89  & 0.97  & 0.99
            &0.77  & 0.87  & 0.89 
            &0.83  & 0.92  & 0.95
\\
&w/o 
            &0.91  & 0.96  & 0.97
            &0.89  & 0.91  & 0.92  
            &0.90  & 0.97  & 0.99
            &0.77  & 0.86  & 0.89 
            &0.83  & 0.91  & 0.95

\\
\hline

\multirow{2}{*}{\textbf{CdMT-Gpt-4o}}
            &w
            &0.93   	&0.97     &0.98 
            &0.88   	&0.91     &0.91 
            &0.96       &0.98     &0.99 
            &0.89   	&0.95     &1.00 
            &0.87   	&0.97     &0.98
\\
&w/o 
            &0.93  & 0.97  & 0.98
            &0.88  & 0.91  & 0.91  
            &0.97  & 0.99  & 0.99
            &0.89  & 0.95  & 1.00 
            &0.85  & 0.96  & 0.97

\\

\bottomrule
\end{tabular}
\label{tab5}
\end{table*}

\subsubsection{\textbf{Multi-step Reasoning}}

After obtaining the context descriptions $\mathcal{D}^i_\text{Cont}$, characteristic descriptions $\mathcal{D}^C_\text{Char}$, and differential descriptions $\mathcal{D}_\text{Diff}$, the LMM performs multi-step reasoning for a target traffic sign. \textbf{Step 1}: The LMM first performs a preliminary understanding of the target traffic sign image based on existing knowledge.
\textbf{Step 2}: The LMM understands the information about the scene around the target traffic sign by referring to the context descriptions. The LMM further narrows the thinking scope by referring to the prior traffic sign hypotheses.
\textbf{Step 3}: By referring to the characteristic descriptions, the LMM understands the basic features of various traffic signs, including shape, color, and composition, and compares the understanding of the target traffic sign image with the characteristic descriptions, thereby stimulating fine-grained TSR.
\textbf{Final}: By referring to the differential descriptions, the LMM gains insights into the differences between the target traffic sign and other similar traffic signs to optimize the recognition results as follows:
\begin{equation}
\mathcal{T}_o^i  = \mathrm{LMM}(\mathcal{I}^i, \mathcal{D}^i_\text{Cont}, \mathcal{D}^C_\text{Char}, \mathcal{D}_\text{Diff}, \mathcal{T}_\text{Multi}),\label{eq8}
\end{equation}
where $\mathcal{T}_\text{Multi}$ represents the designed multi-step prompt, and $\mathcal{T}_o^i$ denotes the final TSR results of the LMM. Through multi-step thinking, the LMM performs feature inference step by step to finally identify the ``real face'' of the target traffic sign. Multi-step thinking can largely stimulate the ability of the LMM to recognize traffic signs at a fine-grained level. Therefore, the fine-grained TSR performance in real-world scenarios of the LMM is improved. Algorithm~\ref{alg:cdmt} outlines the complete process of the proposed CdMT framework.

% \begin{table}[t]
% \centering
% \small
% \caption{Details of each dataset. }
% \setlength{\tabcolsep}{2pt}
% % \begin{adjustbox}{width=0.45\textwidth}
% \begin{tabular}{l|c|c|c|c|c}
% \toprule
% & GTSRB & BTSD & TT-100K & Sapporo& Yokohama \\
% \midrule
% Classes & 43 & 62 & 45 & 19 & 16 \\
% Instances & 200 & 200 & 214 & 90 & 77 \\
% Traffic Sign Extracted & \checkmark &\checkmark & & & \\
% Original Images & & & \checkmark & \checkmark & \checkmark \\
% Thinking-strategy & Two & Two & Multi & Multi & Multi \\
% \bottomrule
% \end{tabular}
% \label{tab1}
% \end{table}
\vspace{-1.2em}
\section{Experiments}
\subsection{Experimental Settings}\label{sub4.1}
 We conducted comprehensive experiments on several datasets, including three benchmark datasets: the German TSR benchmark (GTSRB) dataset~\cite{stallkamp2012man}, the Belgium traffic sign dataset~\cite{mathias2013traffic}, and the Tsinghua-Tencent 100K (TT-100K) dataset~\cite{zhu2016traffic}. TT-100K focuses on complex scenarios in the real world; thus, it is a difficult benchmark to recognize. In addition, to comprehensively evaluate the performance of the proposed method in real-world scenes, we conducted experiments on two Japanese real-world datasets: the Sapporo urban road dataset (Sapporo) and the Yokohama urban road dataset (Yokohama). We perform fine-grained TSR on both open-source and closed-source LLMs. The proposed method does not require model training. \textcolor{black}{However, due to the rate limits of LMM APIs ~\footnote{https://platform.openai.com/account/limits}, we followed the experimental setting strategy in~\cite{yang2023set} and randomly used the subsets of GTSRB, BTSD, and TT-100K validation data in our study. Note that we do not reduce the number of categories in the subset but rather keep it consistent with the categories in the full dataset to comprehensively validate the fine-grained TSR performance of the proposed CdMT framework. To minimize sampling bias, we used stratified random sampling to maintain a balanced class distribution within each subset.} In addition, because the traffic signs in the GTSRB and BTSD datasets have been extracted, multiple thinking is directly performed on them, and because of the lack of original road images, context descriptions are not generated. For the TT-100K, Sapporo, and Yokohama datasets, we use the proposed traffic sign extraction framework to locate and extract traffic signs from the original road images. The common evaluation metric Top-$k$ accuracy, which performs a comprehensive evaluation of TSR performance, was used to evaluate the performance of the proposed fine-grained TSR method.  
Top-$k$ is defined as follows:
\begin{equation} 
{\rm Top\mathchar`-}k = \frac{\mathcal{C}_{k}}{\sum_{i} \mathcal{I}^i}.\label{3}
\end{equation}
Here, \(\mathcal{C}_{k}\) represents the number of correctly recognized target traffic signs in the Top-$k$ results. Considering the challenges of fine-grained TSR in the absence of training data, the Top-$k$ metric can effectively measure the TSR performance. 

\subsection{Experimental Results}\label{sub4.2}

\par Table~\ref{tab2} shows the Top-$k$ fine-grained TSR performance compared with the state-of-the-art methods. We evaluated and validated the proposed method on the three benchmarks and two real-world datasets. As shown in Table~\ref{tab2}, all comparison methods exhibited limited accuracy, reflecting the difficulty of zero-shot fine-grained TSR in the wild. In addition, the recognition performance of the methods of Li et al.~\cite{li2018real} and Zheng et al. (ViT-L)~\cite{zheng2022evaluation} exhibit significant performance differences on datasets from different countries, highlighting that these methods struggle with cross-country TSR in the absence of training data. The Top-$1$ and Top-$3$ accuracies of the proposed method exceed those of the comparison methods on the five datasets with significant improvements compared with the hand-craft feature-based (Song et al.~\cite{yucong2021traffic}, Ren et al.~\cite{ren2009general}), the CNN-based (Gan et al.~\cite{gan2023zero}, DenseNet-121~\cite{huang2017densely}, EfficientNet-B0~\cite{tan2019efficientnet}, Li et al.~\cite{li2018real}), and Transformer-based (Zheng et al. (ViT-L)~\cite{zheng2022evaluation}, Luo et al.~\cite{luo2023pre}) TSR methods, proving the effectiveness of the proposed CdMT framework. \textcolor{black}{We also compare the fine-grained TSR performance of the proposed method with that of advanced transformer architectures (MobileViT~\cite{mehta2021mobilevit}, Swin-Transformer V2~\cite{liu2022swin}, MAE~\cite{he2022masked}, DeiT~\cite{touvron2021training}, CLIP (ViT-B/32)~\cite{radford2021learning},  EVA-02~\cite{fang2023eva}), and cross-domain models (CoOp~\cite{zhou2022learning}, MaPLe~\cite{khattak2023maple}, and CLIP-Adapter~\cite{gao2024clip}). The proposed approach similarly demonstrates promising performance.} In addition, CdMT-enhanced models, including CdMT-Gpt-4v and CdMT-Gpt-4o, exhibit superior performance over baseline LMMs. CdMT-Gpt-4v achieves second-best results across multiple evaluation metrics, whereas CdMT-Gpt-4o consistently leads in terms of Top-1 recognition accuracy on all datasets. This substantial improvement over models such as Gpt-4v and Gpt-4o without CdMT illustrates the ability of the proposed framework to enhance existing LLMs for effective fine-grained TSR. The stability and robustness of the proposed CdMT integration are further evident in the ability of CdMT to maintain high recognition rates across various datasets despite the challenges of cross-country variability in TSR. These results emphasize the potential of our approach in leveraging LMMs for sophisticated and precise TSR applications. 
Furthermore, the latest open-source models, such as LLaVA-NeXT and VITA-1.5, are significantly enhanced by applying the proposed CdMT framework. The CdMT-LLaVA-NeXT and CdMT-VITA-1.5 models demonstrate marked improvements in accuracy across several datasets compared with their original versions. This indicates that the proposed approach not only reinforces closed-source models but also substantially augments the capabilities of open-source models, demonstrating the flexibility and adaptability of the proposed CdMT approach across different types of model architectures.
Note that all experimental results are based on the average of five trials to verify the recognition stability of the proposed method. 
% In addition, we also calculated the standard deviation fluctuation of the proposed method for the five trials on each dataset.
% As shown in Table~\ref{tabst}, the standard deviation of Top-$1$, Top-$3$, and Top-$5$ in the five trials are in a small range for both LMMs. This demonstrates the stable recognition ability of the proposed strategy.

% \begin{table}[h]
% \centering
% \small
% \caption{The Top-$k$ standard deviation of the proposed method. }
% \setlength{\tabcolsep}{0.3pt}
% \renewcommand{\arraystretch}{1.5} % 调整行间距
% % \begin{adjustbox}{width=0.45\textwidth}

% \begin{tabular}{l|c|c|c|c|c|c}
% \toprule
% LMM&Top-k& GTSRB & BTSD & TT-100K & Sapporo& Yokohama \\
% \midrule
% \multirow{3}{*}{\textbf{Gpt-4v}}
% &Top-1 & 0.91$\pm$0.01 & 0.89$\pm$0.01 & 0.90$\pm$0.01 & 0.77$\pm$0.01 & 0.83$\pm$0.02 \\
% &Top-3 & 0.96$\pm$0.01 &  0.91$\pm$0.01 & 0.97$\pm$0.01 & 0.86$\pm$0.02 & 0.91$\pm$0.02\\
% &Top-5& 0.97$\pm$0.01 & 0.92$\pm$0.01 & 0.99$\pm$0.01 & 0.89$\pm$0.01& 0.95$\pm$0.01\\
% \hline
% \multirow{3}{*}{\textbf{Gpt-4o}}
% &Top-1 & 0.93$\pm$0.01 & 0.88$\pm$0.01 & 0.97$\pm$0.01 &0.89$\pm$0.01 & 0.85$\pm$0.01 \\
% &Top-3 & 0.97$\pm$0.01 &  0.91$\pm$0.00 & 0.99$\pm$0.01 & 0.95$\pm$0.01 & 0.96$\pm$0.00\\
% &Top-5& 0.98$\pm$0.01 & 0.91$\pm$0.00 & 0.99$\pm$0.01 & 1.00$\pm$0.01& 0.97$\pm$0.00\\

% \bottomrule
% \end{tabular}
% \label{tabst}
% \end{table}

\par Figure~\ref{fig6} illustrates a recognition example of the proposed method on the Sapporo dataset. We show the detailed prompts and generated descriptions for the proposed CdMT framework. As shown in Fig.~\ref{fig6}, during context description generation, the center coordinates of the target traffic sign are provided in prompts to help the LMM accurately locate the target traffic sign among multiple traffic signs, such as Center Line (910, 271). In addition, the prior traffic sign hypothesis allows the LMM to filter irrelevant answers among all traffic sign candidates. During characteristic description generation, we carefully design the prompts to fully allow LMM to identify key traffic sign features such as shape, color, and composition. The LMM performs in-context learning and generates a brief characteristic description for each traffic sign. By converting the generated descriptions using the LMM's strong recognition of image features, the proposed method reduces the cross-domain discrepancy between the template and target traffic sign images. To generate differential descriptions, similar traffic signs are inserted into the LMM to strengthen its fine-grained recognition capability by emphasizing the differences between similar traffic signs. All input prompts in the proposed method are simple and uniform and do not need to be specially adjusted for different target traffic signs. Figure~\ref{supb5} illustrates recognition examples on the TT-100K dataset. Similar to the Sapporo dataset from Japan, TT-100K is a real-world dataset taken from China. For cross-country traffic signs, the results show that the proposed CdMT framework is general and requires no specific adjustments.

\subsection{Ablation Studies}\label{sub4.3}
\subsubsection{\textbf{Different Thinking Strategies}} To further verify the effectiveness of the proposed multi-step thinking strategy and explore the respective effectiveness of each proposed description. We calculated the Top-$k$ fine-grained TSR performance for different thinking strategies on five datasets, as shown in Table~\ref{tab3}. When no context, characteristic, and differential descriptions exist, target traffic signs are directly input into the LMM for recognition. Table~\ref{tab3} demonstrates that the baseline exhibits the lowest accuracy on all datasets compared with the performance of the thinking strategy. As the number of thinking steps increases, the Top-$k$ TSR recognition accuracy improves, demonstrating the effectiveness of the proposed method. In addition, the results demonstrate that each proposed description improves the fine-grained TSR performance of the LMM. By comparing the results obtained when only one type of description is used, the characteristic description contributes the most to TSR recognition accuracy. Through characteristic descriptions, the LMM can consider the features of the target traffic sign and the descriptions of template traffic signs, thereby improving its fine-grained recognition ability. In addition, context and differential descriptions optimize fine-grained TSR recognition on all five datasets, which is consistent with our hypothesis. 
\par
\begin{figure}[htbp]
       \centering
       \resizebox{0.45\textwidth}{!}{%
           \includegraphics{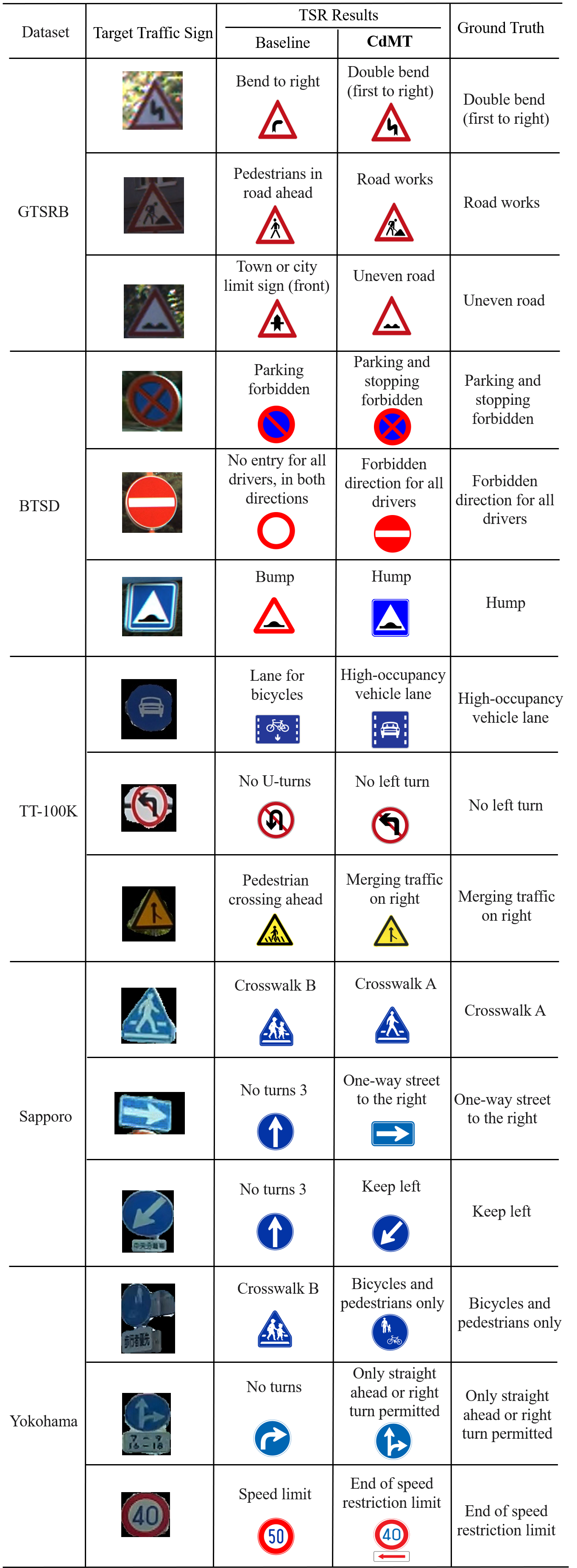}
       }
       \caption{Recognition examples of baseline and proposed CdMT (Gpt-4o).}
       \label{fig9}
   \end{figure}

\begin{figure}[t]
        \centering
        \includegraphics[width=7cm]{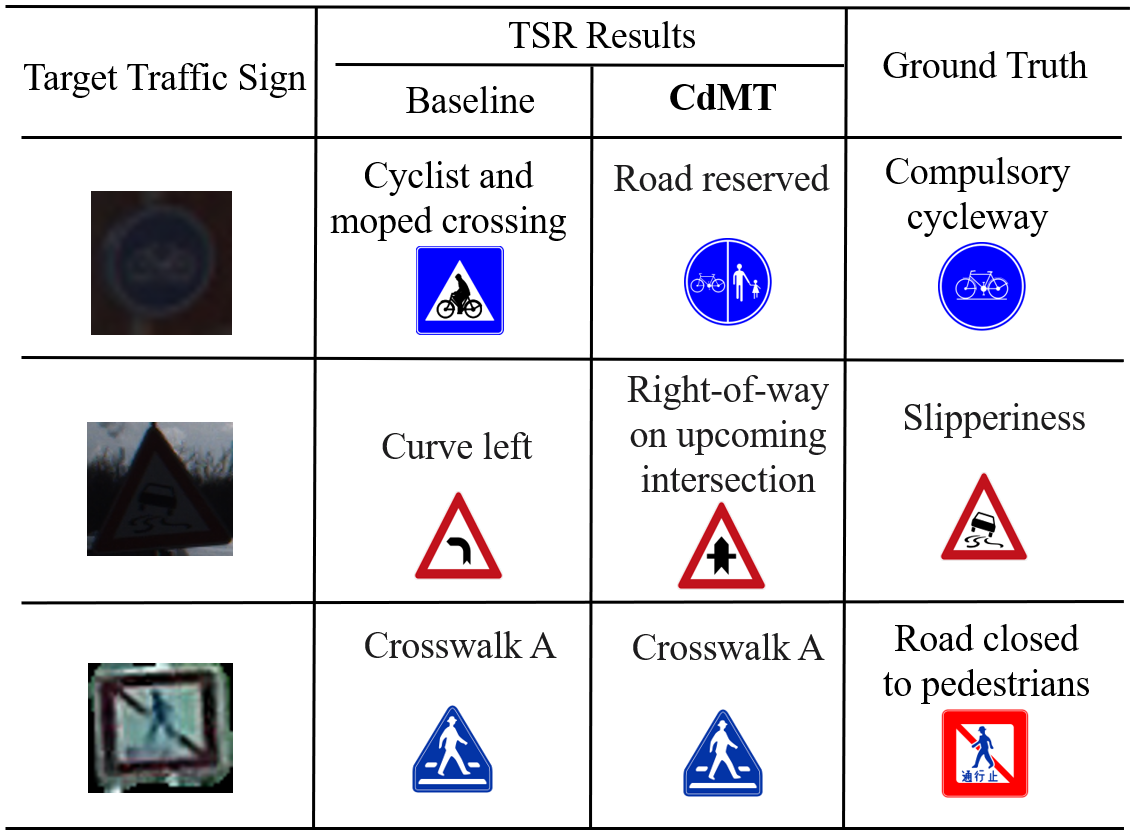}
        \caption{Error recognition examples of baseline and the proposed CdMT (Gpt-4o).}
        \label{fig12}
\end{figure}

 Figure~\ref{fig9} illustrates recognition examples of the baseline (Gpt-4o) and the proposed method on five datasets (CdMT-Gpt-4o). Compared with the baseline, the proposed strategy demonstrates stable recognition performance for traffic signs in real-world scenarios and can be generalized to recognize traffic signs in different countries. In particular, as shown in Fig.~\ref{fig9}, most of the traffic signs identified by the baseline and the proposed strategy exhibit only minor differences. This illustrates that the proposed strategy enables the LMM to fully consider the diversity and similarity of traffic signs for accurate fine-grained level TSR. Figure~\ref{fig12} illustrates examples of recognition errors of the proposed method. When traffic signs are too blurred, understanding the traffic sign images for accurate recognition is difficult.
\subsubsection{\textbf{Hypothesis and Coordinate}} To validate the effectiveness of the proposed prior traffic sign hypothesis and center coordinate prompt optimization method, we experimentally evaluated the effectiveness of different context description generation methods on three real-world datasets, namely, TT-100K, Sapporo, and Yokohama, using the best-performed models. Note that all results in Table \ref{tab4} use context, characteristic, and differential descriptions for multi-step thinking.
As shown in Table~\ref{tab4}, without the prior traffic sign hypothesis and center coordinate prompt optimization, i.e., with only simple image background descriptions in the contextual description, the Top-$k$ fine-grained TSR performance is reasonably similar to the accuracy presented in Table~\ref{tab3} obtained using only the characteristic and differential descriptions. Because the characteristic and differential descriptions are provided, only simple background descriptions of images in the context description contribute to improving the fine-grained capability of the LMM. The situation is also similar when only the center coordinates optimization is performed. Although the LMM can locate the target traffic sign from multiple traffic signs in the original road image and simply describe the features, the simple descriptions in the contextual description contribute little because characteristic descriptions are already provided. When only the prior traffic sign hypothesis is used without center coordinate prompt optimization, the LMM struggles to locate the target traffic sign from multiple traffic signs in the original road image, thereby generating confusing descriptions. The confusing descriptions negatively affect accuracy. When both the prior traffic sign hypothesis and the center ordinate prompt optimization are performed, the Top-$k$ fine-grained TSR performance is improved by locating the target traffic signs and filtering irrelevant answers.

\begin{figure}[t]
        \centering
        \includegraphics[width=8cm]{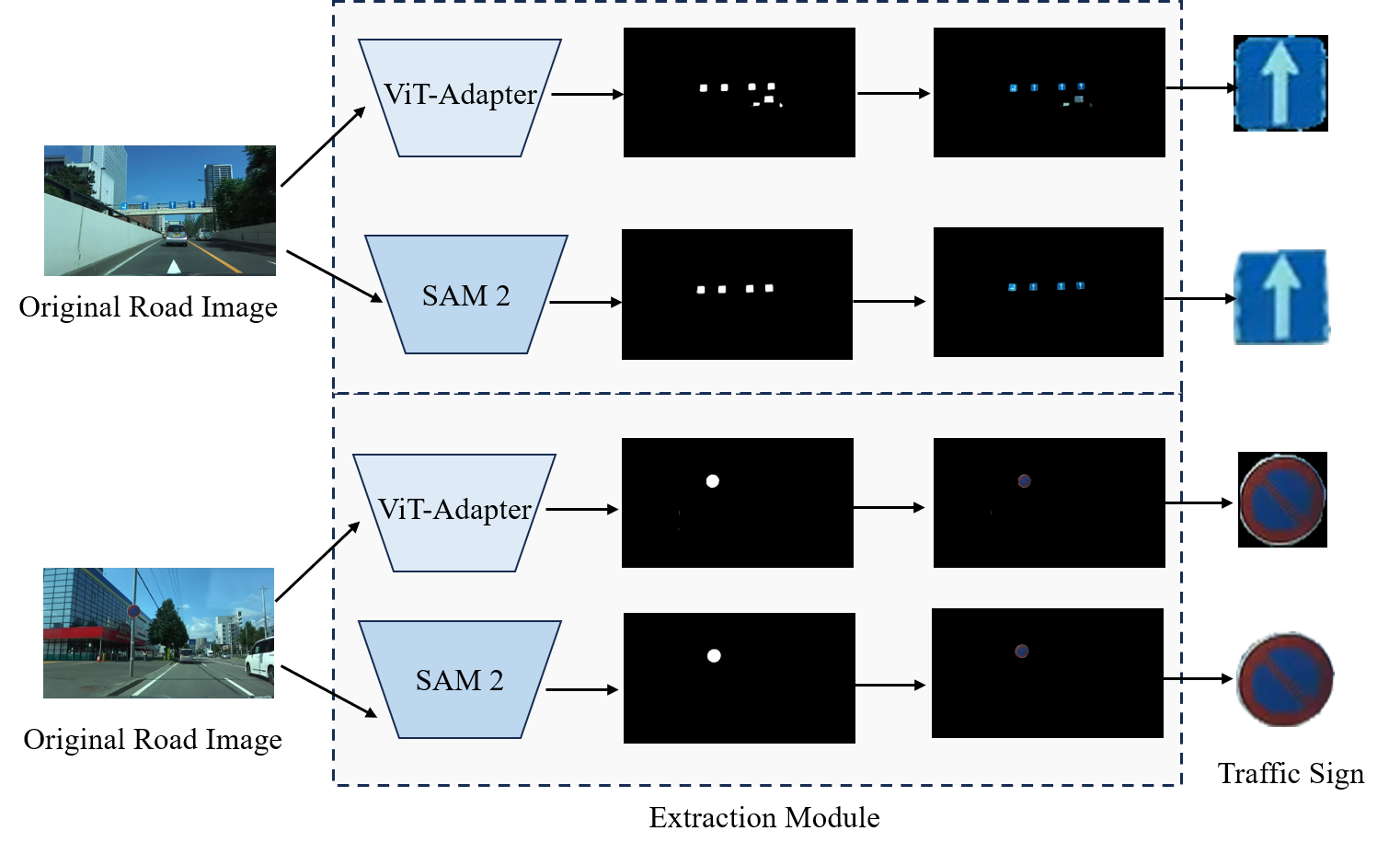}
        \caption{Examples of traffic sign extraction using the designed extraction module under different segmentation models.}
        \label{fig10}
\end{figure}

% \begin{table}[t]
% \centering
% \small
% \caption{Segmentation IoU of different extraction modules on the TT-100K subset.}
% \begin{tabular}{lcc}
% \toprule
% \textbf{Segmentation Model} & \textbf{Mean IoU} \\
% \midrule
% SAM 2          & 0.96 \\
% ViT-Adapter    & 0.89 \\
% \bottomrule
% \end{tabular}
% \label{tab:seg_iou}
% \end{table}

\begin{table}[tbp]
\centering
\small
\setlength{\tabcolsep}{1pt} % 可适当变化调整列间距
\renewcommand{\arraystretch}{1.2} % 行高

\caption{\textcolor{black}{TSR results of CdMT-Gpt-4o under different extraction modules.}}
\label{tab:different_seg}
\begin{tabular}{l ccc ccc ccc}
\toprule
\multirow{2}{*}{\textbf{Model}} & \multicolumn{3}{c}{\textbf{TT-100K}} & \multicolumn{3}{c}{\textbf{Sapporo}} & \multicolumn{3}{c}{\textbf{Yokohama}} \\
\cmidrule(r){2-4} \cmidrule(r){5-7} \cmidrule(l){8-10}
               & Top-1 & Top-3 & Top-5                & Top-1 & Top-3 & Top-5                & Top-1 & Top-3 & Top-5 \\
\midrule
ViT-Adapter    
& 0.93  & 0.96  & 0.98                 
& 0.84  & 0.91  & 0.96                
& 0.80  & 0.88  & 0.93  \\
SAM 2          & 0.97  & 0.99  & 0.99                 & 0.89  & 0.95  & 1.00                 & 0.85  & 0.96  & 0.97  \\
\bottomrule
\end{tabular}
\end{table}

\subsubsection{\textbf{Thinking Orders}} Table~\ref{tab5} compares TSR performance for different numbers of thinking steps. For the GTSRB and BTSD datasets, we change the order of thinking for characteristic and differential descriptions. For the TT-100K, Sapporo, and Yokohama datasets, we change the thinking order for context and characteristic descriptions. The experimental results are presented in Table~\ref{tab5}. After the order of thinking is changed, TSR performance remains almost the same as the initial performance, demonstrating the robustness of the proposed method. The absence of changes in the cues obtained by the LMM, even when the order of thinking is changed, indicates no significant difference in recognition accuracy.

\subsubsection{\textbf{Extensibility}} The previous experiments demonstrated that the proposed multi-step thinking strategy could be easily extended to both open- and closed-source LMMs, such as CdMT-LLaVA-NeXT, CdMT-VITA-1.5, CdMT-Gpt-4v, and CdMT-Gpt-4o, and maintains robust performance. In addition, in the designed traffic sign extraction module, the segmentation model is not limited to a specific model and can easily be extended to advanced models. \textcolor{black} {Table~\ref{tab:different_seg} presents the TSR performance of CdMT-Gpt-4o based on different extraction modules on the TT-100K, Sapporo, and Yokohama datasets. With segment anything model 2 (SAM 2)~\cite{ravi2024sam2} as the extraction module, the model consistently achieves higher Top-1, Top-3, and Top-5 accuracies across all datasets than when ViT-Adapter~\cite{chen2023vision} is used. These results demonstrate that the extraction module benefits from advances in segmentation models, because stronger segmentation yields more accurate and reliable traffic sign extraction, which directly improves overall TSR performance.} Figure~\ref{fig10} illustrates traffic sign extraction examples with SAM 2 and ViT-Adapter. As shown in Fig.~\ref{fig10}, under different segmentation models, target traffic signs are extracted using the designed extraction module. The most advanced segmentation model SAM 2 performs better extraction on traffic signs.

\subsubsection{\textbf{Inference Speed}}  Table~\ref{tab6} presents the inference speed of the proposed method based on different LMMs and segmentation approaches. The integration of ViT-Adapter-base for segmentation yields an inference time of approximately 0.4 s per road image, indicating substantial efficiency. In contrast, employing the SAM 2-base extraction module improves this performance, achieving real-time extraction capabilities. Notably, among the LMMs evaluated, the Gpt-4o with the proposed CdMT framework achieves the fastest comprehensive inference, with a total time of 1.2 s per traffic sign. In addition, the CdMT variant based on the latest open-source model VITA-1.5 achieves an inference time of 1.9 s per traffic sign. The modularity of the proposed framework allows its seamless extension to future  LMM variations, suggesting that enhancements in model architectures and computational strategies can be swiftly integrated, potentially further reducing inference times.
\begin{table}[t]
\small
    \centering
    \caption{Inference speed for each traffic sign. ``s'' represents seconds}
    \begin{tabular}{c|c|c}
    \toprule
    Extraction & LMM & Inference Speed \\
    \hline
    \multirow{5}{*}{\centering ViT-Adapter-base} &
    LLaVA-v1.5 & 6.4s \\
    & LLaVA-NeXT & 6.0s \\
    & VITA-1.5 & 2.3s \\
    & Gpt-4v & 2.0s \\
    & Gpt-4o & 1.6s \\
    \hline
    \multirow{5}{*}{\centering SAM 2-base} &
    LLaVA-v1.5 & 6.0s \\
    & LLaVA-NeXT & 5.6s \\
    & VITA-1.5 & 1.9s \\
    & Gpt-4v & 1.6s \\
    & Gpt-4o & 1.2s \\
    \bottomrule
    \end{tabular}
\label{tab6}
\end{table}

\begin{figure}[ht]
        \centering
        \includegraphics[width=8cm]{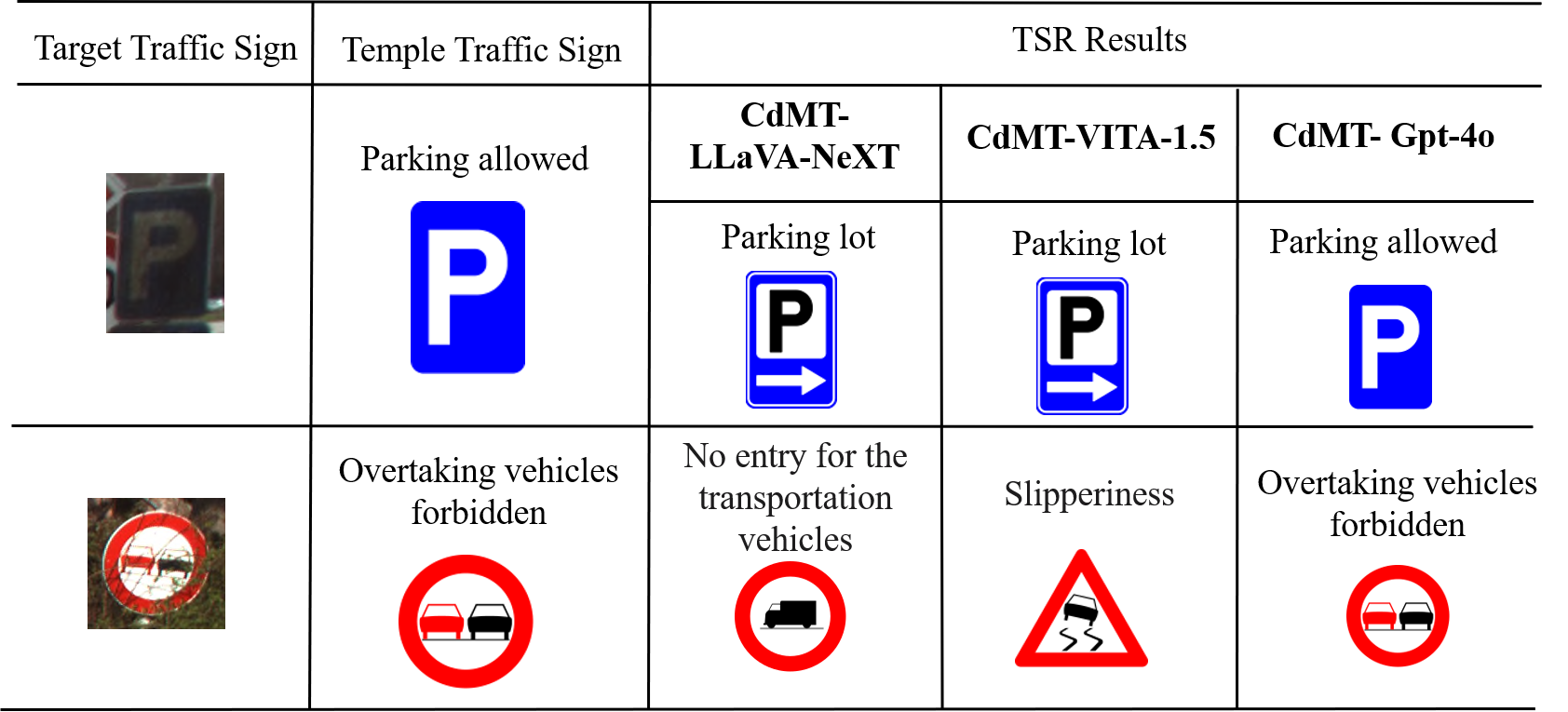}
        \caption{CdMT recognition results for significant domain shift samples.}
        \label{fig:domain_shift}
\end{figure}

\subsubsection{Significant Domain Shift}

~\textcolor{black}{To further explore the performance of CdMT under significant domain shifts, we conducted a case study involving two challenging real-world samples, as shown in Fig.~\ref{fig:domain_shift}. The first sample, ``Parking allowed,'' suffers from a background color shift that causes significant deviations from the template traffic sign. The second sample, ``Overtaking vehicles forbidden,'' is partially occluded by tree leaves, further increasing the recognition difficulty. Both samples represent difficulties encountered in real-world TSR. We evaluated the TSR performance of three CdMT variants, each employing a different LMM: CdMT-LLaVA-NeXT, CdMT-VITA-1.5, and CdMT-Gpt-4o. The results demonstrate that both CdMT-LLaVA-NeXT and CdMT-VITA-1.5 misclassified the ``Parking allowed'' sign as the visually and semantically similar ``Parking lot.'' Similarly, these models misidentified the ``Overtaking vehicles forbidden.'' In contrast, CdMT-Gpt-4o correctly recognized both samples, demonstrating greater robustness to significant domain shifts. The results highlight the critical importance of underlying LMM capabilities in the presence of significant domain shifts.}

\begin{table*}[tbp]
\centering
\small

\setlength{\tabcolsep}{1.5pt} % 调整列间距
\renewcommand{\arraystretch}{1.2} % 调整行间距
\caption{\textcolor{black}{TSR results of CdMT-Gpt-4o under different characteristic description lengths. }}
\begin{tabular}{c|c|ccc|ccc|ccc|ccc|ccc}
\toprule
\multirow{2}{*}{\textbf{LMM}}&\multirow{2}{*}{\textbf{Description Length}}&\multicolumn{3}{c|}{\textbf{GTSRB}} 
& \multicolumn{3}{c|}{\textbf{BTSD}} 
& \multicolumn{3}{c|}{\textbf{TT-100K}}
& \multicolumn{3}{c|}{\textbf{Sapporo}} 
& \multicolumn{3}{c}{\textbf{Yokohama}}
\\
%\midrule
\cline{3-17}
 &&\textbf{Top-1} & \textbf{Top-3} & \textbf{Top-5} & \textbf{Top-1} & \textbf{Top-3} & \textbf{Top-5} & \textbf{Top-1} & \textbf{Top-3} & \textbf{Top-5} & \textbf{Top-1} & \textbf{Top-3} & \textbf{Top-5} &\textbf{Top-1} & \textbf{Top-3} & \textbf{Top-5}\\
\midrule

\multirow{2}{*}{\textbf{CdMT-Gpt-4o}}
            &Short
            &0.91   	&0.94     &0.97 
            &0.86   	&0.89     &0.89 
            &0.90       &0.94     &0.96 
            &0.86   	&0.92     &0.98 
            &0.75   	&0.82     &0.91
\\
&Medium 
            &0.93  & 0.97  & 0.98
            &0.88  & 0.91  & 0.91  
            &0.97  & 0.99  & 0.99
            &0.89  & 0.95  & 1.00 
            &0.85  & 0.96  & 0.97

\\
&Long 
            &0.92  & 0.97  & 0.98
            &0.88  & 0.91  & 0.92  
            &0.97  & 0.98  & 0.99
            &0.89  & 0.94  & 0.99 
            &0.85  & 0.95  & 0.96

\\

\bottomrule
\end{tabular}
\label{tab_length}
\end{table*}
\subsubsection{Description Length}

\begin{figure}[t]
        \centering
        \includegraphics[width=9cm]{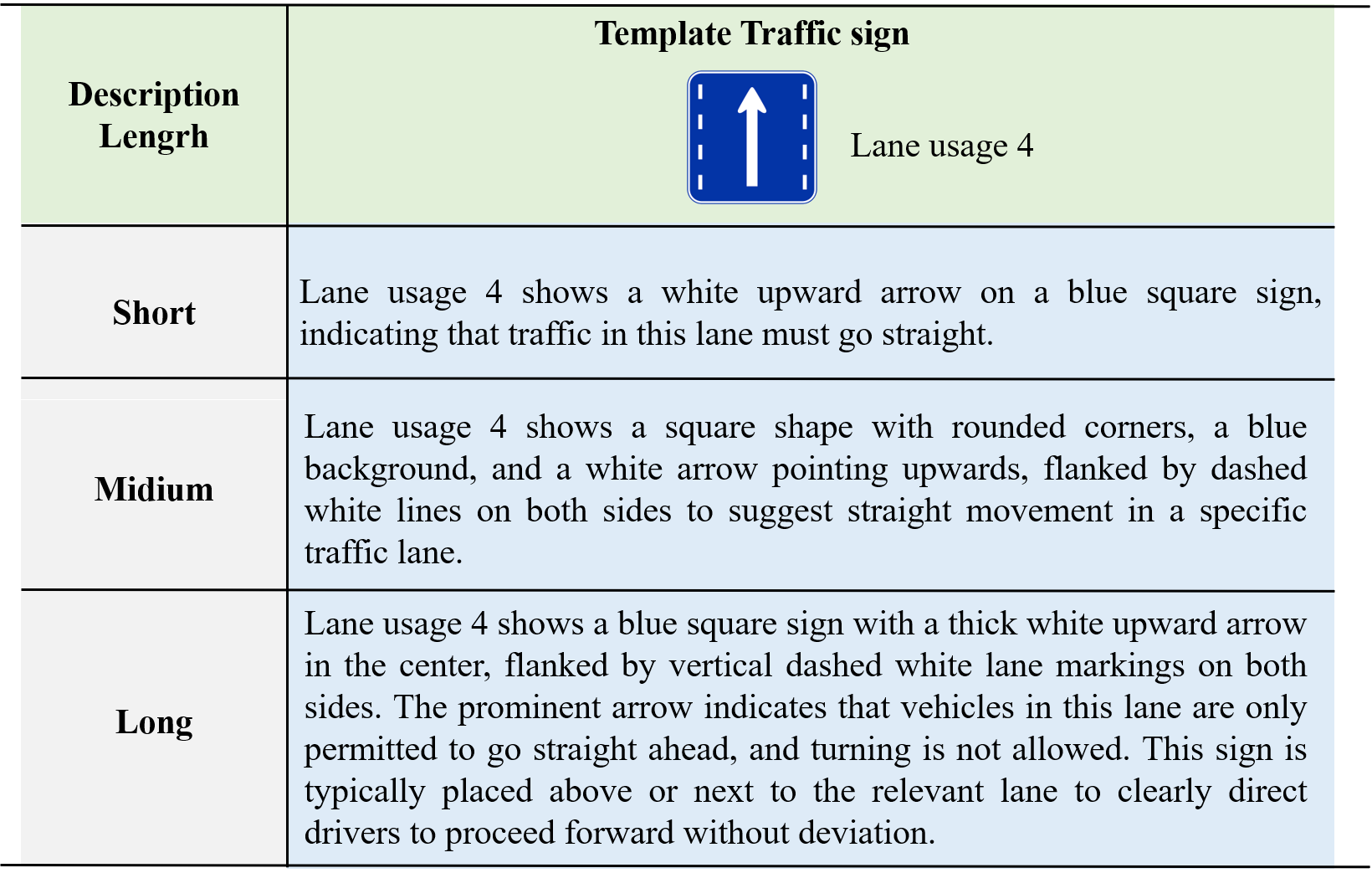}
        \caption{\textcolor{black}{Examples of short, medium, and long characteristic descriptions.}}
        \label{fig:length}
\end{figure}

\textcolor{black}{To evaluate the effect of characteristic description length on TSR performance, we conducted an ablation study with CdMT-Gpt-4o. The results are summarized in Table~\ref{tab_length}. Three settings were evaluated: short, medium, and long descriptions. Examples are illustrated in Fig.~\ref{fig:length}. Across all datasets, the medium and long descriptions yield consistently higher Top-1, Top-3, and Top-5 accuracies. Notably, the medium setting achieves similar performance to the long setting while requiring less computational cost. In contrast, short descriptions lead to a clear drop in performance on all five datasets, likely due to insufficient representation of fine-grained visual features. These results demonstrate that overly short descriptions may fail to capture key discriminative features required for cross-domain TSR, whereas providing more detailed descriptions does not lead to further improvements. In general, an appropriate description length is important to maximize accuracy while maintaining computational efficiency.}

\section{Discussion}
\subsection{Test Set Contamination}
LMMs are trained on large amounts of internet data; thus, there are concerns and speculation that they have memorized public benchmarks~\cite{oren2023proving}. In this study, we not only tested our method on three public benchmark datasets (GTSRB, BTSD, TT-100K) but also on two private datasets (Sapporo and Yokohama). Our method exhibits consistent and robust performance on all five datasets. The two private datasets could not have been used in model training. Therefore, test set contamination does not exist in the proposed method.
\subsection{Importance and Application}
The proposed method can achieve efficient TSR in natural dynamic road environments and maintain stable TSR performance in different countries without the need for training data. This highlights its significant application value. Collecting and preparing data for training and testing across various countries is costly, especially given differing data and privacy policies and the challenges in obtaining data from less developed regions. By reducing the need for extensive data collection, our approach not only reduces costs but also promotes equity. Current advanced driving assistance systems and autonomous driving technologies are typically limited to certain regions, neglecting less developed areas. By achieving effective cross-country TSR, the proposed method can extend existing technologies to underserved regions, thereby promoting greater equity.

\subsection{Limitation}
\subsubsection{Determination of Similar Traffic Signs}
\textcolor{black}{In this study, we designed differential descriptions for LMMs and demonstrated the effectiveness of these descriptions. However, similar traffic signs are selected based on expert knowledge to generate these descriptions, which may introduce subjectivity and limit scalability to larger or more diverse traffic sign databases. In future work, we plan to investigate automatic methods for determining similar traffic signs to improve consistency and enable broader applicability of the proposed framework.}
\subsubsection{Performance under Different Weather Conditions} \textcolor{black}{In this study, five datasets, including three public datasets (GTSRB, BTSD, TT-100K) and two private datasets (Sapporo, Yokohama), were used to verify the performance of the proposed method. However, all five datasets were collected under sunny weather. Thus, the traffic sign images are relatively clear. Under weather conditions such as rain, fog, and snow, traffic sign images may be blurred, which affects TSR performance. Improving TSR performance under such conditions is a direction we look forward to exploring in the future. For example, future work may explore designing the thinking process with weather-specific context cues to improve recognition robustness under adverse weather.}
\begin{table}[t]
\small
\centering
\caption{\textcolor{black}{Computation time cost of CdMT-Gpt-4o. Here, $N_C$ represents the class number of template traffic signs; $N_D$ represents the number of similar traffic signs; ``Phase Type'' indicates whether each phase is performed online during inference or offline as a preprocessing step. All time costs are in s.}}
\begin{tabular}{lcc}
    \toprule
    CdMT Phase & Time Cost (s) & Phase Type \\ 
    \midrule
    Traffic Sign Extraction   & 0.1                  & Online \\
    Context Description       & 0.4                  & Online \\
    Characteristic Description & $0.3 \times N_C$     & Offline \\
    Differential Description  & $0.3 \times N_D$     & Offline \\
    Multi-step Reasoning      & 0.7                  & Online \\
    \bottomrule
\end{tabular}
\label{tabcomputation_time}
\end{table}
\subsubsection{Computational Complexity and Latency}
\textcolor{black}{Table~\ref{tabcomputation_time} summarizes the computation time cost of each phase in the proposed CdMT-Gpt-4o framework. The traffic sign extraction, context description, and multi-step reasoning phases require 0.1, 0.4, and 0.7 s, respectively. Notably, the characteristic and differential description generation phases can be performed offline, and the results cached; thus, these computations are required only once. Although the online execution of multi-step reasoning is crucial for the effectiveness of the proposed approach, it may also introduce additional computational overhead and inference latency, which can present challenges during deployment in real-time or resource-constrained environments. Addressing these concerns may involve applying model distillation techniques to compress LMMs into more compact and efficient models, thereby substantially reducing the inference time while maintaining accuracy. In addition, optimizing the multi-step reasoning pipeline by removing redundant operations or incorporating adaptive reasoning based on input complexity will allow for more efficient inference tailored to specific scenarios. These improvements will further enhance the practicality of the proposed CdMT framework for latency-sensitive, real-world applications.}

\subsubsection{Incorporation of Standardized Sign Taxonomies}

\textcolor{black}{Although our current approach generates characteristic descriptions via in-context learning based on template traffic signs, we acknowledge that standardized sign taxonomies, such as those provided by the Vienna Convention or global, interpretable rules for traffic signs defined by ISO 3864, can be considered. Integrating such standardized taxonomies into our CdMT prompt design can improve both the interpretability and consistency of the generated characteristic descriptions and improve model generalization across different domains. In future work, we plan to explore the incorporation of formalized sign taxonomy information into the prompt strategy.}

\section{Conclusion}
\textcolor{Black}{In this study, we proposed the CdMT framework for constructing a general fine-grained TSR method. The proposed framework is simple, effective, and easily extensible. The designed multi-thinking strategy stimulates the zero-shot fine-grained recognition ability of LMMs for traffic signs. The results of the experiments conducted on three benchmark datasets and two real-world datasets demonstrate the effectiveness of the proposed method. Future work will focus on developing automatic methods for identifying similar traffic signs, improving robustness under varying weather conditions, and further enhancing computational efficiency to facilitate real-time deployment in practical scenarios.}
\section*{Acknowledgments}
Some data in this study were provided by Japan Radio Co., Ltd. This study was supported in part by
JSPS KAKENHI Grant Numbers JP23K21676, JP23K11141, JP23K11211, JP24K02942, JP24K23849, JP25K21218 and JST BOOST, Japan Grant Number JPMJBS2426.
\bibliographystyle{elsarticle-num}
\bibliography{KBS}
\end{document}